%% file: main.tex
\documentclass[10pt,twocolumn,letterpaper]{article}

\usepackage[pagenumbers]{cvpr}      %

\input{preamble}

\definecolor{cvprblue}{rgb}{0.21,0.49,0.74}
\usepackage[pagebackref,breaklinks,colorlinks,allcolors=cvprblue]{hyperref}

\def\confName{CVPR}
\def\confYear{2027}

\title{Can Visual World Models See Force?}

\author{
\quad Felix Tristram$^{1,2}$
\quad Benjamin D. Killeen$^{1,2}$
\quad Nassir Navab$^{1,2}$\\
$^1$ Technical University of Munich \quad $^2$ Munich Center for Machine Learning
}

\begin{document}
\maketitle
\input{sec/0_abstract}    
\input{sec/1_intro}
\input{sec/2_related_work}
\input{sec/3_method}
\input{sec/4_experiments}
\input{sec/5_conclusion}
\clearpage
{
    \small
    \bibliographystyle{ieeenat_fullname}
    \bibliography{main}
}

\input{sec/X_suppl}

\end{document}

%% file: preamble.tex
\usepackage{amsmath,amssymb,booktabs}
\usepackage[T1]{fontenc}

\newcommand{\forcese}[1]{\scalebox{0.55}{$\pm#1$}}

\usepackage{placeins}

%% file: sec/0_abstract.tex
\begin{abstract}
World models and vision-language-action policy models are increasingly proposed as foundations for robot manipulation. They predict how scenes evolve and which actions to take, yet successful manipulation depends not solely on following observable trajectories but also on the hidden variable of interaction force. Neurophysiology has shown that humans possess a visual prior for this hidden variable, motivating us to ask the question: Do the representations of visual foundation models also capture the forces involved in an interaction? We define force awareness as the ability of representations to (1) indicate when contact force is applied, (2) track how it changes during an interaction, and (3) reflect how large it is. We probe frozen image, video, world- and policy model representations with a lightweight readout trained on synchronized video and force measurements from two manipulation datasets. Our experiments show that most models capture when contact occurs but not how much force is being applied. Surprisingly, robot post-training does not improve this either, highlighting an important gap in current models capabilities.
\end{abstract}

%% file: sec/1_intro.tex
\section{Introduction}
\label{sec:intro}

World-action models may enable generalist robot manipulation by learning to predict how the world changes together with the actions a robot takes. DreamZero, for example, jointly predicts future video and actions, building on large-scale video pretraining~\cite{ye2026world}. Related approaches use predictive video representations for planning or train vision-language-action models across diverse robot demonstrations~\cite{assran2025v,kim2024openvla,black2024pi_0,bjorck2025gr00t}. These efforts raise the prospect that physical understanding can emerge from more video, larger models, and more robot data. Yet inaccurate force application has been identified as a failure mode of robot foundation models~\cite{fact2026,zhang2025ta}, which commonly predict motion targets while leaving force control to low-level controllers~\cite{assran2025v,kim2024openvla,black2024pi_0,bjorck2025gr00t,agarwal2026cosmos}. These failures raise the question of whether the models' representations capture the forces involved in manipulation, beyond the motion they predict.

An example of similar motion trajectories but different force application can be seen in everyday activities such as opening a fridge vs. a cabinet door. The doors can follow similar trajectories but require different forces because of friction or resistance that is difficult to see. Vision can provide useful cues here: humans use visual appearance to anticipate an object's weight and scale their initial lifting forces~\cite{buckingham2009living}, while visual feedback during lifting helps correct expectation-driven force errors~\cite{buckingham2010lifting}. These findings motivate us to investigate whether visual world models also acquire representations that implicitly encode interaction force.

\begin{figure}[t]
\centering
\includegraphics[width=0.8\linewidth]{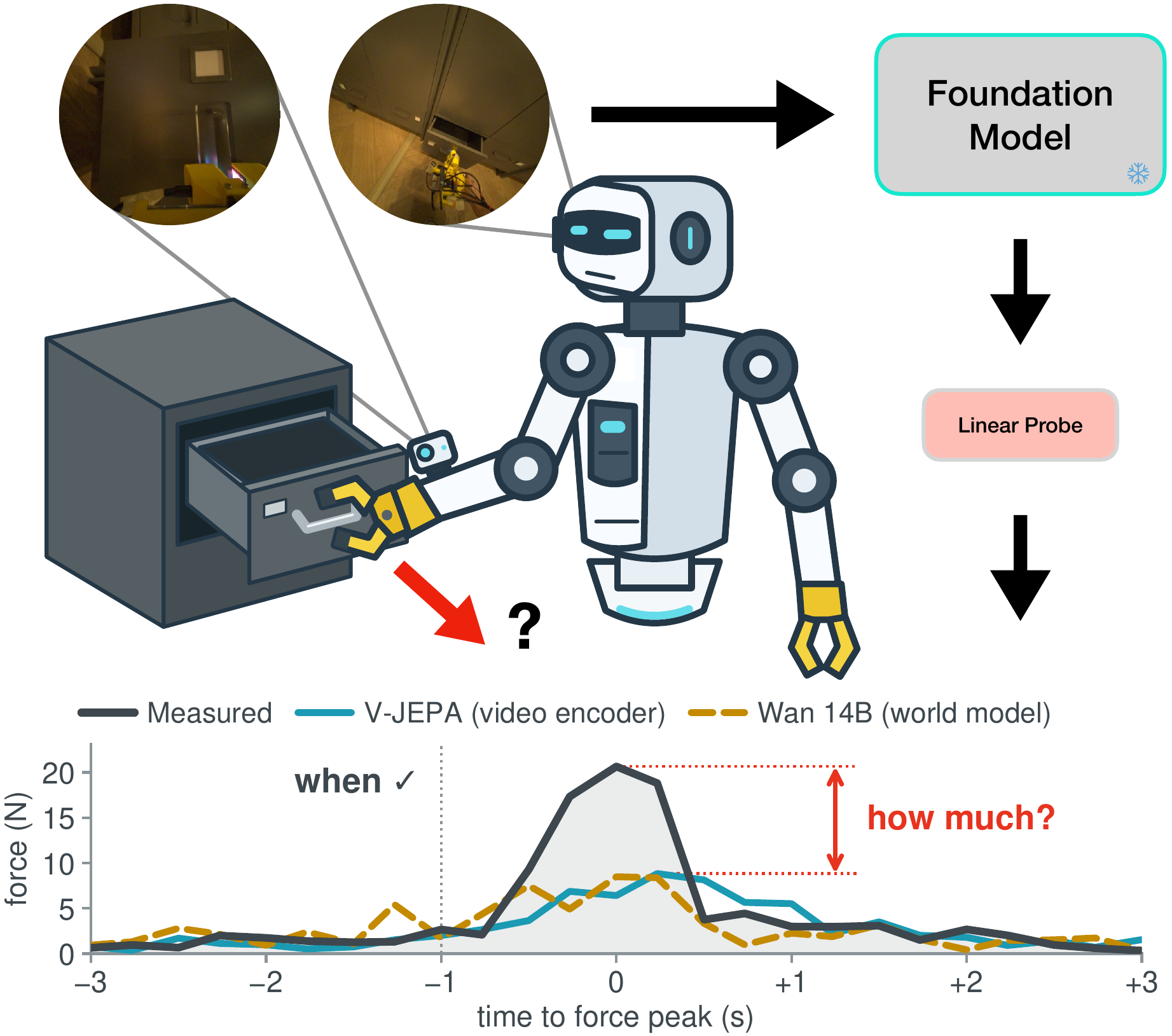}
\caption{Does a world model know how much force it takes to open a cabinet? We train a lightweight probe on frozen features to decode measured force (top). Bottom: probe predictions for an example Hoi!\ gripper-view contact. Both models respond when contact occurs, but neither recovers how much force is applied.}
\label{fig:teaser}
\vspace{-6mm}
\end{figure}

Prior evaluations of visual foundation models reveal difficulties with hidden mechanical properties and broader physical understanding~\cite{tung2023physion++,chow2025physbench}. At the same time, probing studies find geometric information, physical plausibility signals, and motion variables in pretrained representations~\cite{el2024probing,huang2026much, esmati2026invisible}. Readouts of elasticity, viscosity, and friction provide further evidence of accessible physical information, although accurate value prediction in real-world videos remains challenging~\cite{zhan2026inferring}.
Together, these studies establish the systematic evaluation of physical and geometric information in pretrained representations as a research contribution in its own right. We extend this line of inquiry to interaction force by asking: how much force can be decoded from models trained for general visual, generative, and action objectives?

We probe 13 different models spanning image and geometry encoders, self-supervised video encoders, generative world models, and vision-language and policy models, on synchronized video and force measurements from the Hoi! dataset~\cite{engelbracht2026hoi}, and the larger robotic RH20T~\cite{fang2024rh20t} dataset. The model set contains matched comparisons: base checkpoints and their robot-trained counterparts, different sizes within the Wan and Cosmos3 families, and two camera views of the same interactions. We separate three capabilities that are often conflated: detecting \emph{when} contact force is applied, tracking \emph{how} it changes within an interaction, and recovering \emph{how much} force is applied. Our main findings are:

\begin{itemize}
\item \textbf{Detecting contact is easy, recovering magnitude is not.} Most representations detect when force is applied, including single-image encoders, but not how much. Nearly all compress the measured force range, and on Hoi!\ most are less accurate than predicting the mean contact force.
\item \textbf{V-JEPA is the exception.} In the Hoi!\ gripper view and on RH20T, it tracks force best and explains the most variance in contact force, although at most $17\%$. Neither video input nor model size accounts for this: VideoMAEv2 observes the same clips, and world models of up to 14B parameters remain below the 1B-parameter V-JEPA encoder.
\item \textbf{Robot training does not reliably improve force awareness.} Of four robot-trained world and VLA models, only DreamZero tracks force better than the model it builds on, and only on Hoi!. On RH20T it does not, although its predictions cover more of the measured force range.
\end{itemize}
\vspace{0.5pt}\noindent Together, these results suggest that force awareness does not come for free with video generation or robot training. It can be learned though, through direct force supervision: fine-tuning V-JEPA on RH20T force labels improves force tracking there and transfers to Hoi!\ without retraining. This makes force supervision a promising direction, and force a quantity worth evaluating alongside geometry, video quality and task success.

%% file: sec/2_related_work.tex
\section{Related Work}
\label{sec:rel_work}

\paragraph{World models and policies for manipulation.}
Video models learn how scenes change through different training objectives. VideoMAEv2 reconstructs masked video content~\cite{wang2023videomae}, V-JEPA predicts missing features~\cite{mur2026v}, and LeVJEPA learns consistent features across views~\cite{kuhn2026levjepa}, while Wan and Cosmos learn to generate video~\cite{wan2025wan,nvidia2025cosmospredict2,agarwal2026cosmos}. These models increasingly serve robot control. V-JEPA~2 plans actions in its feature space~\cite{assran2025v}, and world-action models such as DreamZero and the DROID checkpoints of Cosmos~3 predict future video and robot actions jointly~\cite{ye2026world,agarwal2026cosmos}. VLA models take a different route and add action prediction to a vision-language model, as in OpenVLA, $\pi_0$, and GR00T, which builds on Cosmos-Reason2~\cite{kim2024openvla,black2024pi_0,bjorck2025gr00t,grootcard,reason2card}. We probe models from all of these groups, with DINOv3 and VGGT-$\Omega$ as image and geometry baselines~\cite{simeoni2025dinov3,wang2026vggt}. None of them receives force as input or predicts it directly.

\paragraph{Why force matters for manipulation.}
Contact-rich tasks remain difficult for these approaches. FACT~\cite{fact2026} traces the failures of VLA models in such tasks to two sources, imprecise actions and missing responses to contact force, and TA-VLA~\cite{zhang2025ta} shows that giving a VLA model torque as input and as a prediction target improves its success in contact-rich tasks. Other works add physical feedback around the policy. Lee et al.~\cite{lee2019making} combine vision, force/torque, and robot state to learn representations for insertion, Reactive Diffusion Policy pairs a slow visual policy with fast tactile feedback~\cite{xue2025reactive}, and FP2~\cite{fp2_2026} separates task-level action generation from force regulation driven by wrench and proprioceptive history. FD-VLA~\cite{fdvla2026} instead distills measured force into a representation conditioned on vision and robot state, so that the deployed policy needs no force sensor. This relies on force being partly predictable from what the robot sees. The feedback itself comes from dedicated force sensors, as in RH20T and Hoi!~\cite{fang2024rh20t,engelbracht2026hoi}, or from optical tactile sensors such as DenseTact~2.0~\cite{do2022densetact}.

\paragraph{Estimating force from visual observations.}
Force can also be estimated from images directly, and recent benchmarks have made this a task of its own~\cite{engelbracht2026hoi}. Pham et al.~\cite{pham2015towards} combine RGB-D tracking with mechanical constraints to infer manipulation forces, accounting for the different forces that can produce the same motion. Learning-based methods exploit visual cues at contact. VisGel predicts tactile observations from external images~\cite{li2019connecting}, and PressureVision and EgoPressure estimate hand pressure from images with synchronized pressure measurements~\cite{grady2022pressurevision,zhao2025egopressure}. EgoPHI estimates 3D hand-object contact and force from egocentric images and object geometry, supervised with simulated forces~\cite{ilic2026egophi}, and Hoi!\ provides multi-view video with measured forces during articulated manipulation~\cite{engelbracht2026hoi}. All of these methods are designed and supervised for force. They show that force is visible to some extent, but not whether a model trained without it captures it.

\paragraph{Probing physical understanding.}
What a pretrained model has learned can be measured by training a small readout on its frozen features. Such probes show that visual features predict surface geometry but struggle to match surfaces across viewpoints~\cite{el2024probing}, that image, video, and 3D encoders have different strengths in scene understanding~\cite{man2024lexicon3d}, and that video models encode depth, 3D points, and camera pose~\cite{huang2026much}. For physics, the picture is less favorable. Physion tests whether models anticipate object contact~\cite{bear2021physion}, Physion++ shows that they do not reliably infer latent mechanical properties from observed interactions~\cite{tung2023physion++}, and Kang et al.~\cite{kang2024far} find that video generators fail to apply physical laws beyond their training distribution. Vision-language models show similar weaknesses across physical properties and dynamics~\cite{chow2025physbench}, and when asked for numerical estimates of size, velocity, and acceleration, they give plausible but inaccurate answers~\cite{puyin2026quantiphy}. Some physical information is nevertheless present inside pretrained models. V-JEPA separates physically possible from impossible videos, unlike models that predict pixels~\cite{garrido2025intuitive}. Probing studies show that physical plausibility and motion variables are decodable from video-diffusion states~\cite{esmati2026invisible}, and lightweight readouts of video features estimate elasticity, viscosity, and friction, although less accurately than cue-based reference methods~\cite{zhan2026inferring}. Force has so far entered these models only as an input. Force Prompting~\cite{gillman2026force} and PhyCo~\cite{narayanan2026phyco} train video generators, the latter on top of Cosmos-Predict2, to follow a specified force. Whether the features of pretrained models describe the force in an observed interaction has not been measured. We address this with a probe on synchronized video and force measurements, and separate detecting contact, tracking force changes, and recovering force magnitude.

%% file: sec/3_method.tex
\section{Probing Force in Frozen Representations}
\label{sec:method}

\begin{figure*}[t]
\centering
\includegraphics[width=\textwidth]{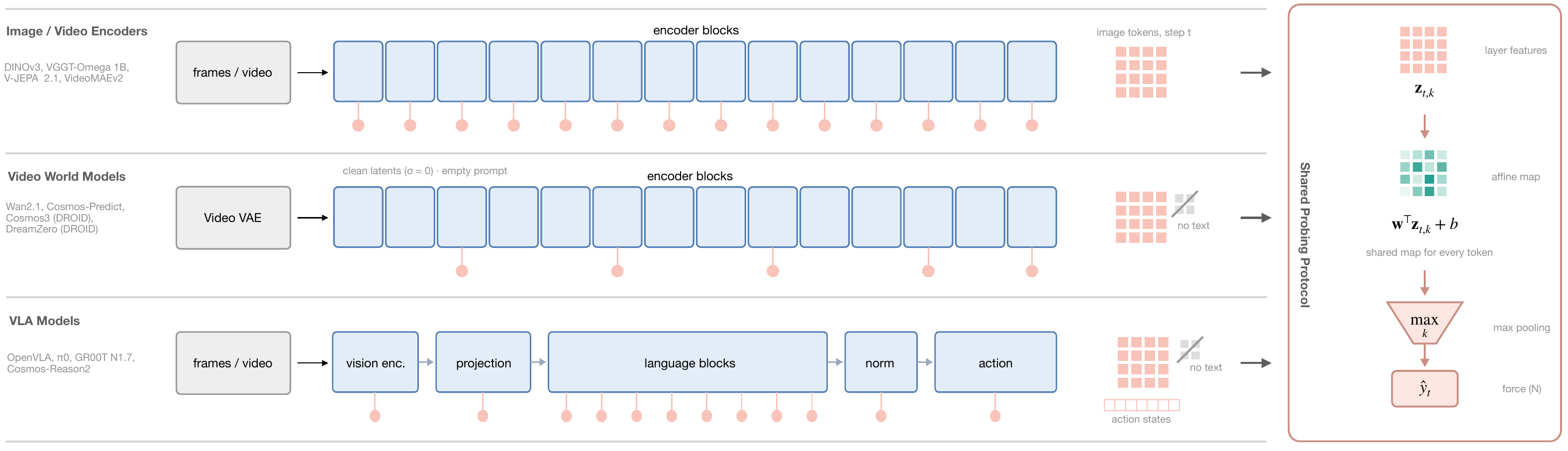}
\caption{\textbf{Probing protocol.} We extract spatial tokens from intermediate layers of frozen image and video encoders, video world models, and VLA models. The same readout is trained on every feature source: a shared affine map scores each token, and max pooling over the tokens of a time step gives the predicted force.}
\label{fig:method}
\end{figure*}

We measure how much interaction force can be decoded from a frozen foundation model. Following prior probing works~\cite{el2024probing}, we keep all model weights fixed and train only a lightweight readout on its features (Fig.~\ref{fig:method}). The readout is identical for all models; only the feature extraction differs between model families due to architectural differences.

\subsection{Force decoding task}
\label{sec:task}
The input is a video window of a manipulation with synchronized force measurements $\mathbf{F}(\tau)\in\mathbb{R}^3$. We divide the window into $T$ time steps of two consecutive frames and regress the mean force magnitude of each step,
\begin{equation}
 y_t = \frac{1}{|B_t|}\sum_{\tau\in B_t}\|\mathbf{F}(\tau)\|_2,
 \label{eq:target}
\end{equation}
where $B_t$ contains the force samples recorded during step $t$. Windows cover the full recording, including periods without contact. We regress the magnitude rather than the force vector because it is independent of the sensor's coordinate frame and the direction of motion.

\subsection{Probe}
\label{sec:readout}
For a model $m$ and extraction location $\ell$, let $\mathbf{z}_{t,k}\in\mathbb{R}^{d}$ be the feature of spatial token $k$ at time step $t$. The probe applies one affine map to every token and max-pools the resulting scalars over the tokens $\mathcal{K}$ of the time step:
\begin{equation}
 \widehat{u}_t = \max_{k\in\mathcal{K}}
 \left(\mathbf{w}^{\top}\mathbf{z}_{t,k}+b\right).
 \label{eq:probe}
\end{equation}
The probe has $d+1$ parameters, no hidden layers, and no temporal module. Max pooling lets it select the tokens that carry contact information without object or contact annotations. We train a separate probe for every location $\ell$.

Targets are standardized with the mean $\mu$ and standard deviation $\sigma$ of the training forces, and the probe minimizes
\begin{equation}
 \mathcal{L}
 = \frac{1}{T}\sum_{t=1}^{T}
 \rho\!\left(\widehat{u}_{t}-\frac{y_{t}-\mu}{\sigma}\right),
 \label{eq:loss}
\end{equation}
averaged over the windows of a batch, where $\rho$ is the smooth-$L_1$ loss. Half of each batch is drawn from the training windows with the highest peak force (top quartile) and half uniformly from all windows. At test time, the predicted force in newtons is
\begin{equation}
 \widehat{y}_t=\operatorname{softplus}\!\left(\sigma\,\widehat{u}_t+\mu\right),
 \label{eq:output_transform}
\end{equation}
which is nonnegative.

\subsection{Feature extraction}
\label{sec:representations}
We probe intermediate layers of each model, as the final layer is often not the most informative one. All models are run in a single forward pass on the observed video. Extracted tokens are arranged on a $T\times H\times W$ grid; for models with a finer temporal resolution, we average features over each time step. The supplement lists all extraction details.

\paragraph{Image and video encoders.}
For V-JEPA~2.1~\cite{mur2026v} and VideoMAEv2~\cite{wang2023videomae}, we probe the patch tokens after every encoder block. Following their released inference protocols, V-JEPA processes the full window ($T=64$), and VideoMAEv2 consecutive 16-frame clips. DINOv3~\cite{simeoni2025dinov3} encodes each frame independently, and we probe its patch tokens after every block. For VGGT-$\Omega$~\cite{wang2026vggt}, we probe the concatenated frame- and global-attention patch tokens of every aggregator block.

\paragraph{Video world models.}
For Wan~2.1~\cite{wan2025wan}, Cosmos-Predict2~\cite{nvidia2025cosmospredict2}, Cosmos3~\cite{agarwal2026cosmos}, and DreamZero~\cite{ye2026world}, we encode the observed video with the model's video autoencoder and pass the clean latents through the diffusion transformer at zero noise with an empty text prompt (following \cite{el2024probing, huang2026much}). We probe the video tokens after every transformer block for Wan~1.3B, Cosmos-Predict2, and Cosmos3-Edge, and at five evenly spaced depths for the larger Wan I2V~14B, Cosmos3-Nano, and DreamZero. Image-to-video models are conditioned on the first frame. For a clean comparison, DreamZero and the Cosmos3 DROID receive the video stream only and no robot state or actions.

\paragraph{VLA models.}
OpenVLA~\cite{kim2024openvla}, $\pi_0$~\cite{black2024pi_0}, GR00T~N1.7~\cite{grootcard}, and Cosmos-Reason2~\cite{reason2card}, the vision-language model that GR00T builds on, receive the image at each time step together with a text instruction (Sec.~\ref{sec:datasets}). We probe the image tokens at each stage of the model: after the vision encoder, after the projection into the language model, after every language-model block, and after its final normalization. GR00T additionally provides the output of its vision-language adapter, and OpenVLA the hidden states of its seven action tokens, which then form the pooling set $\mathcal{K}$. OpenVLA and $\pi_0$ observe a single image. GR00T uses its native two-frame history, and Cosmos-Reason2 its native video input.

\subsection{Evaluation}
\label{sec:evaluation}
\paragraph{Metrics.}
We evaluate the three aspects of force awareness separately. Contact episodes $E$ are segmented from the measured force with a 2\,N threshold (details in the supplement); $\mathcal{E}$ is the set of all episodes and $\mathcal{C}$ the set of all time steps inside them.

\emph{When is force applied?} We label each time step as contact if its measured force is at least 2\,N and use the predicted force $\widehat{y}_t$ as the detection score. We report the area under the ROC curve (AUROC), which equals the probability that a randomly chosen contact step receives a higher predicted force than a randomly chosen non-contact step. It is $0.5$ for a random prediction and $1$ for perfect separation. AUROC depends only on the ordering of the predictions, so it measures whether the probe responds to contact, regardless of the predicted force values.

\emph{How does force change?} We report the mean Pearson correlation between predicted and measured force within each contact episode,
\begin{equation}
 r=\frac{1}{|\mathcal{E}|}\sum_{E\in\mathcal{E}}
 \operatorname{corr}\!\left(\widehat{\mathbf{y}}_E,\mathbf{y}_E\right),
 \label{eq:correlation}
\end{equation}
which is invariant to the scale and offset of the prediction. We also report the peak hit rate (Hit), the share of episodes in which the highest predicted force lies within one time step of the measured peak.

\emph{How much force is applied?} We report the coefficient of determination over all contact time steps,
\begin{equation}
 R^2=1-\frac{\sum_{t\in\mathcal{C}}(y_t-\widehat{y}_t)^2}
 {\sum_{t\in\mathcal{C}}(y_t-\overline{y}_{\mathcal{C}})^2},
 \label{eq:magnitude_metrics}
\end{equation}
where $\overline{y}_{\mathcal{C}}$ is the mean contact force. Unlike $r$, $R^2$ requires correct force values, and a negative $R^2$ means that the probe is less accurate than predicting this mean. Additionally, to measure whether the prediction covers the measured force range, we report the response amplitude
\begin{equation}
 A=\frac{\sum_{E\in\mathcal{E}}\operatorname{std}(\widehat{\mathbf{y}}_E)}{\sum_{E\in\mathcal{E}}\operatorname{std}(\mathbf{y}_E)},
 \label{eq:amplitude}
\end{equation}
the ratio of predicted to measured force variation within an episode. $A=1$ means the prediction varies as much as the measured force, and $A<1$ that it is compressed. Finally, $\mathrm{MAE}_{90+}$ is the mean absolute error on time steps in the top decile of training forces.

\paragraph{Layer selection.}
All selection uses the validation scenes. For each location, we keep the training epoch $e^*_\ell$ with the highest validation $R^2$. We then report each metric at the location that maximizes it on validation,
\begin{equation}
 \ell^*=\arg\max_{\ell}R^2_{\mathrm{val}}(\ell,e^*_{\ell}),\quad
 \ell^*_r=\arg\max_{\ell}r_{\mathrm{val}}(\ell,e^*_{\ell}),
 \label{eq:selection}
\end{equation}
with $R^2$ reported at $\ell^*$ and all other metrics at $\ell^*_r$. Standard errors are computed with a bootstrap over scenes (see suppl.).

%% file: sec/4_experiments.tex
\section{Experiments}
\label{sec:experiments}

\subsection{Datasets}
\label{sec:datasets}

\paragraph{Hoi!.}
Hoi!~\cite{engelbracht2026hoi} records interactions with articulated objects such as drawers, doors, and cabinets, performed by a human with an instrumented gripper that measures the applied force. We use the instrumented-gripper sessions and evaluate their two synchronized camera views, \emph{human} and \emph{gripper}. We follow the official split of the benchmark, which holds out six environments for evaluation, and train on the remaining 12 locations, so that no evaluation scene is seen during training. For selection, we divide the six evaluation environments into three for validation and three for testing (724, 109, and 151 windows for training, validation, and testing). The evaluation environments contain 131 human-view and 132 gripper-view contact episodes.

\paragraph{RH20T.}
RH20T~\cite{fang2024rh20t} contains teleoperated robot manipulation across more than 140 tasks with a force-torque sensor at the wrist. We use the in-hand camera (mirroring \textit{gripper} from Hoi!) and split by operator-task group. With 7,241 training windows and 11,768 test windows from 2,439 recordings, it is ten times larger than Hoi!\ in training data, far more diverse in objects and motions, and its almost 80 times larger test-set allows for statistically meaningful analysis. As a video-free reference on RH20T, we report a task prior that predicts the average training force of the recorded task without observing the video (supplement).

On both datasets, a window covers 64 frames at 7.5\,Hz (8.5\,s) and is divided into $T=32$ time steps. Video world models receive the same window at their native frame rate and an empty text prompt on both datasets (following \cite{el2024probing, huang2026much}). Models that require an instruction receive the fixed instruction ``Manipulate the drawer.'' on Hoi!; on RH20T, they receive the task description of each recording. Statistics for all splits are given in the supplement.

\subsection{Implementation details}
\label{sec:implementation}
We train one probe per model, view, and extraction location for 20 epochs with AdamW, learning rate $10^{-3}$, weight decay $10^{-2}$, and a batch size of eight windows. Epochs and locations are selected on the validation split only (Sec.~\ref{sec:evaluation}). We report all metrics on the pooled validation and test splits, neither of which is used to train the probe; on Hoi!, these are the six official evaluation environments. Test-only tables are in the supplement. Contact episodes are segmented with a 2\,N threshold. Values after $\pm$ are standard errors from a bootstrap that resamples whole scenes on Hoi!\ and operator-task groups on RH20T. The supplement describes segmentation and bootstrap in detail.

\input{tables/joint_selection/table_main}

\subsection{Main findings}
\label{sec:findings}
We report force decoding on Hoi!\ in Table~\ref{tab:force_best} and on RH20T in Table~\ref{tab:rh20t}, and add the amplitude, peak, and high-force diagnostics for Hoi!\ in Table~\ref{tab:force_diagnostics}. One observation is clear above all others: force is poorly decodable from every model we probe. The best probe reaches a within-episode correlation of only $0.53$ on Hoi!\ and $0.20$ on RH20T, and none explains more than $17\%$ of the variance in contact force. The differences between models discussed below lie within this low range, and we read them against the bootstrap standard errors reported with every score.

\input{figures/qualitative/figure_qualitative}

\paragraph{Detecting contact is easy, recovering magnitude is not.}
Every probe detects contact well above chance on both datasets (Table~\ref{tab:rh20t} and supplement). This requires no temporal input: in the Hoi!\ gripper view, the single-image encoders detect contact as reliably as V-JEPA. How much force is applied is a different matter. On Hoi!, contact $R^2$ is negative for every model in the human view and for all but three in the gripper view (Table~\ref{tab:force_best}), so most probes are less accurate than predicting the mean contact force. Table~\ref{tab:force_diagnostics} shows what goes wrong. Nearly all probes predict a smaller force variation than was measured ($A<1$), often less than half of it, and they fail where force is highest: on the top decile of forces, even the best probes are off by more than 6\,N, only about $40\%$ less than a constant prediction. Timing is recovered better than magnitude, but not reliably. Most models place the force peak within one time step in fewer than $40\%$ of the episodes, where a random time step achieves $22\%$. In Figure~\ref{fig:qualitative}, the predictions rise when the gripper makes contact but stay far below the measured peak, and no model follows a sustained push. Detecting contact is therefore weak evidence that a model understands force (Spearman $\rho=0.32$ between AUROC and $r$ in the Hoi!\ gripper view).

\input{tables/rh20t/table_rh20t}

\paragraph{V-JEPA is the exception.}
V-JEPA~2.1 has the highest correlation and the highest contact $R^2$ in the Hoi!\ gripper view and on RH20T, and it is the only model whose contact $R^2$ on Hoi!\ lies more than two standard errors above zero. The diagnostics agree: it locates the force peak more often than any other model in both Hoi!\ views and on RH20T, and it has the lowest high-force error in the human view and on RH20T. Video input alone does not explain this. VideoMAEv2 observes the same clips and tracks force within one standard error of V-JEPA on both datasets, yet it does not recover magnitude, with lower $R^2$ and higher MAE$_{90+}$ on RH20T. Model size does not explain it either, since world models with up to 14B parameters remain below the 1B-parameter V-JEPA encoder. While V-JEPA is the best of the models we probe, it is not a solution: in the human view, its $R^2$ is negative like that of every other model.

\paragraph{Robot training does not reliably improve force awareness.}
One could expect models trained on robot data to encode force better, since they have seen many manipulation episodes. Our model set contains four pairs of a base checkpoint and its robot-trained counterpart, and in three of them robot training makes no measurable difference. The DROID versions of Cosmos3-Edge and Cosmos3-Nano track force within one standard error of their base models in both Hoi!\ views, as does Edge DROID on RH20T. Likewise, GR00T stays within one standard error of Cosmos-Reason2, the vision-language model it builds on (supplement). The fourth pair is mixed: DreamZero tracks force better than its base model Wan I2V~14B in the Hoi!\ gripper view ($0.46$ vs.\ $0.35$, more than two standard errors apart) but not on RH20T ($0.15$ vs.\ $0.18$). None of the robot-trained models reaches a positive contact $R^2$ on Hoi!.

\paragraph{The findings hold on RH20T.}
Hoi!\ is small, and its standard errors of contact $R^2$ are often as large as the differences between models. RH20T tests the same findings on almost 80 times more test windows, and the standard error of contact $R^2$ is two to seven times smaller for every model (Table~\ref{tab:rh20t}). The findings discussed above do not change. Probes detect contact but compress the force range, V-JEPA leads in tracking, magnitude, peak localization, and high-force error, and no robot-trained model tracks force better than its base model. With ten times more training data, contact $R^2$ turns positive for most models but stays below $0.1$ for all of them except V-JEPA. RH20T also separates the families more clearly: image encoders fall well behind all video models, and VLA models remain last. Figure~\ref{fig:overlay_rh20t} shows two test recordings with overlays of the tokens that each probe responds to. In a), opening a microwave, the models respond well and the probe has the highest response on the gripper itself. In recording b), pressing the button of a power strip, none of the models estimate force well and token response is focused on the background as well as the gripper. We suspect that the forward motion of the interaction might be what is picked up in this sequence.

\input{tables/joint_selection/table_force_diagnostics}

\subsection{Analysis by model family}
\label{sec:families}

\noindent\textbf{Image and geometry encoders.}
DINOv3 and VGGT-$\Omega$ track gripper-view force on Hoi!\ about as well as most video models (Table~\ref{tab:force_best}). This is largely an effect of contact onset and release: restricted to time steps above 2\,N, their correlation drops to $0.13$ and $0.18$ (supplement). RH20T confirms this picture. On its more diverse tasks, both reach only $r\approx0.08$ and locate the force peak in a quarter of the episodes, compared with more than a third for the video encoders. Per-frame features thus reveal that contact takes place, but not how force develops during it.

\noindent\textbf{Video encoders.}
V-JEPA and VideoMAEv2 are both trained by masked video modeling and differ mainly in what they predict: VideoMAEv2 reconstructs pixels, whereas V-JEPA predicts features. Both track force, but only V-JEPA recovers part of its magnitude. We hypothesize that predicting in feature space favors learning representations that abstract away from the appearance and dynamics individually by merging both, which makes unseen variables such as force more accessible. Similarly, Garrido et al.~\cite{garrido2025intuitive} find that feature-predicting video models detect physically impossible events, whereas pixel-predicting models do not.

\input{figures/qualitative/figure_overlay_rh20t}

\noindent\textbf{Video world models.}
World models are trained to predict how a scene evolves, yet on Hoi!\ they track force no better than single-image encoders. Neither scale nor depth helps: Wan~14B is on par with Wan~1.3B, and correlation is nearly flat across transformer blocks, whereas the video encoders build up force information with depth (supplement). DreamZero is the interesting case. Its predictions cover far more of the measured force range than those of its base model Wan I2V~14B ($A=0.82$ vs.\ $0.47$ in the Hoi!\ gripper view, $0.69$ vs.\ $0.53$ on RH20T), on par with V-JEPA (Figs.~\ref{fig:qualitative}a and~\ref{fig:overlay_rh20t}a), and it has the lowest high-force error among the world models on both datasets, while the DROID checkpoints of Cosmos3 show no such gain. We speculate that this is an effect of jointly predicting future video and robot actions: the video features must then carry what determines the next action, which may include how forceful an interaction is. This remains a hypothesis though, since DreamZero differs from Wans architecture by processing video chunks causally. 

\noindent\textbf{VLA models.}
VLA models are the weakest family. OpenVLA and $\pi_0$ stay well below single-image DINOv3 in the Hoi!\ gripper view, and on RH20T $\pi_0$ and GR00T barely track force at all and predict less than a fifth of the measured force variation (Table~\ref{tab:rh20t}). They do detect contact, so what is missing is how force evolves and how large it is. We see two likely reasons. First, their training signal never requires force: a demonstrated motion can be imitated from what is observable, without representing the force behind it. Second, their vision encoders are trained for alignment with language, which rewards recognizing what is in the scene rather than subtle cues of contact. Indeed, the features that leave the vision encoder are already weak, and the language model adds little (supplement).

\input{tables/lora_outlook/table_lora_outlook}

\subsection{Learning force from data}
\label{sec:lora}
None of the models above is trained with force, and none encodes it well. Does a model that is supervised with force encode it better, and does what it learns transfer to another dataset? To test this, we fine-tune V-JEPA~2.1 with LoRA adapters on the RH20T force labels and probe its final block (Table~\ref{tab:lora_outlook}). On RH20T itself, both tracking and magnitude improve substantially ($r$ from $0.19$ to $0.33$, contact $R^2$ from $0.10$ to $0.23$). More interestingly, the adapted encoder transfers to Hoi!. When we apply the RH20T-trained readout without any retraining, it tracks gripper-view force with $r=0.21$, whereas the same readout on the frozen encoder does not ($0.03$). What transfers is the temporal profile and not the magnitude. Contact $R^2$ is strongly negative for both encoders, because the readout keeps the force normalization of RH20T, whose forces are measured with a different sensor on a different robot. Unlike $r$, $R^2$ penalizes such errors in scale and offset. A newly trained Hoi!\ probe removes this mismatch: the adapted final block reaches $r=0.52$ compared with $0.43$ for the frozen one, while the gain in contact $R^2$ is within its standard error. Results on the human view do not change in either setting. Force supervision on one robot dataset thus improves force tracking on another, which robot training without force did not achieve in any of our model pairs. The adapted model is still far from accurate, but force supervision appears to be a more direct route to force-aware representations than more video or more demonstrations.

%% file: tables/joint_selection/table_main.tex
\begin{table}[t]
\centering
\caption{Force decoding on the \textbf{Hoi! Dataset} with scene-cluster bootstrap standard errors. We report how well each model \textit{tracks} force changes ($r$) and if they can estimate \textit{how much} force is applied (contact $R^2$), sorting rows by best gripper-view $r$ for each model family. Video Encoders outperform all other methods, followed not far behind by generative world models.}
\label{tab:force_best}
\footnotesize
\setlength{\tabcolsep}{1.8pt}
\resizebox{\columnwidth}{!}{%
\begin{tabular}{@{}lcccc@{}}
\toprule
& \multicolumn{2}{c}{Human view} & \multicolumn{2}{c}{Gripper view} \\
\cmidrule(lr){2-3}\cmidrule(lr){4-5}
Representation & $r\uparrow$ & $R^2\uparrow$ & $r\uparrow$ & $R^2\uparrow$ \\
\midrule
\multicolumn{5}{@{}l}{\textit{Image and geometry encoders}} \\
DINOv3 ViT-H+ & $0.173\forcese{0.039}$ & $-0.033\forcese{0.045}$ & $0.406\forcese{0.027}$ & $0.008\forcese{0.054}$ \\
VGGT-Omega 1B & $0.237\forcese{0.022}$ & $-0.047\forcese{0.041}$ & $0.400\forcese{0.034}$ & $0.018\forcese{0.049}$ \\
\midrule
\multicolumn{5}{@{}l}{\textit{Video encoders}} \\
V-JEPA 2.1 ViT-G & $0.337\forcese{0.048}$ & $-0.055\forcese{0.043}$ & $0.527\forcese{0.019}$ & $0.166\forcese{0.061}$ \\
VideoMAEv2 giant & $0.348\forcese{0.027}$ & $-0.037\forcese{0.050}$ & $0.480\forcese{0.040}$ & $-0.012\forcese{0.067}$ \\
\midrule
\multicolumn{5}{@{}l}{\textit{Generative world models}} \\
DreamZero-DROID & $0.279\forcese{0.040}$ & $-0.069\forcese{0.091}$ & $0.458\forcese{0.037}$ & $-0.118\forcese{0.123}$ \\
Cosmos3-Nano & $0.114\forcese{0.057}$ & $-0.297\forcese{0.056}$ & $0.419\forcese{0.029}$ & $-0.037\forcese{0.084}$ \\
Cosmos3-Nano DROID & $0.084\forcese{0.030}$ & $-0.263\forcese{0.058}$ & $0.383\forcese{0.053}$ & $-0.063\forcese{0.116}$ \\
Cosmos-Predict2 2B & $0.159\forcese{0.036}$ & $-0.168\forcese{0.054}$ & $0.368\forcese{0.045}$ & $-0.146\forcese{0.095}$ \\
Cosmos3-Edge & $0.110\forcese{0.056}$ & $-0.192\forcese{0.050}$ & $0.358\forcese{0.051}$ & $-0.204\forcese{0.076}$ \\
Wan 2.1 I2V 14B & $0.225\forcese{0.036}$ & $-0.097\forcese{0.042}$ & $0.353\forcese{0.023}$ & $-0.064\forcese{0.070}$ \\
Cosmos3-Edge DROID & $0.071\forcese{0.066}$ & $-0.215\forcese{0.033}$ & $0.338\forcese{0.047}$ & $-0.086\forcese{0.040}$ \\
Wan 2.1 1.3B & $0.250\forcese{0.049}$ & $-0.177\forcese{0.058}$ & $0.333\forcese{0.034}$ & $-0.047\forcese{0.066}$ \\
\midrule
\multicolumn{5}{@{}l}{\textit{Vision--language / action models}} \\
Cosmos-Reason2 (video) & $0.155\forcese{0.049}$ & $-0.125\forcese{0.071}$ & $0.327\forcese{0.046}$ & $-0.117\forcese{0.058}$ \\
GR00T N1.7 (2 frames) & $0.137\forcese{0.041}$ & $-0.087\forcese{0.081}$ & $0.300\forcese{0.027}$ & $-0.129\forcese{0.068}$ \\
$\pi_0$ & $0.150\forcese{0.030}$ & $-0.213\forcese{0.093}$ & $0.231\forcese{0.044}$ & $-0.098\forcese{0.041}$ \\
OpenVLA-7B & $0.094\forcese{0.010}$ & $-0.080\forcese{0.078}$ & $0.152\forcese{0.024}$ & $-0.061\forcese{0.053}$ \\
\bottomrule
\end{tabular}%
}
\end{table}

%% file: figures/qualitative/figure_qualitative.tex
\begin{figure*}[t]
\centering
\includegraphics[width=\textwidth]{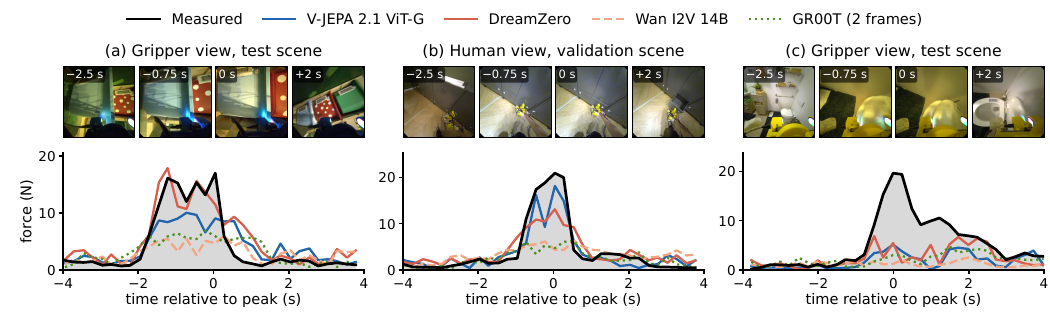}
\caption{Qualitative comparison of force timelines from the Hoi! dataset around three contact peaks for selected models. Frames are labeled with their time relative to the peak. (a) DreamZero follows the gripper-view force profile, V-JEPA responds at lower amplitude, and DreamZero's base model Wan I2V 14B and GR00T remain nearly flat. (b) V-JEPA approaches a 21\,N human-view peak. (c) A sustained push that no model follows.}
\label{fig:qualitative}
\end{figure*}

%% file: tables/rh20t/table_rh20t.tex
\begin{table}[t]
\centering
\caption{Force decoding on the \textbf{RH20T} dataset, with operator-task group bootstrap standard errors. We group rows by model family and sort by $r$ within each. We report force tracking $r$, magnitude $R^2$, contact detection AUROC, peak force hit-rate, standard deviation $A$ and MAE$_{90+}$. Generative world models perform closer to video encoders than on Hoi!. VLAs have extremely compressed force ranges $A$ and overall poor force tracking performance. VJEPA2.1 and DreamZero are the only models with $A$ higher than $0.53$ and the best and second best MAE respectively.}
\label{tab:rh20t}
\footnotesize
\setlength{\tabcolsep}{1.8pt}
\resizebox{\columnwidth}{!}{%
\begin{tabular}{@{}lcccccc@{}}
\toprule
Representation & $r\uparrow$ & $R^2\uparrow$ & AUROC$\uparrow$ & A & Hit (\%)$\uparrow$ & MAE$_{90+}\downarrow$ \\
\midrule
\multicolumn{7}{@{}l}{\textit{Video-free reference}} \\
Task prior (best of 3) & $0.023\forcese{0.005}$ & $-0.207\forcese{0.127}$ & $0.678\forcese{0.017}$ & $0.29$ & $16.7$ & $14.09$ \\
\midrule
\multicolumn{7}{@{}l}{\textit{Image and geometry encoders}} \\
DINOv3 ViT-H+ & $0.081\forcese{0.008}$ & $0.037\forcese{0.013}$ & $0.748\forcese{0.014}$ & $0.36$ & $26.3$ & $10.87$ \\
VGGT-Omega 1B & $0.077\forcese{0.008}$ & $0.058\forcese{0.012}$ & $0.714\forcese{0.012}$ & $0.31$ & $26.2$ & $12.41$ \\
\midrule
\multicolumn{7}{@{}l}{\textit{Video encoders}} \\
V-JEPA 2.1 ViT-G & $0.203\forcese{0.019}$ & $0.133\forcese{0.032}$ & $0.755\forcese{0.011}$ & $0.71$ & $36.4$ & $9.62$ \\
VideoMAEv2 giant & $0.192\forcese{0.019}$ & $0.071\forcese{0.021}$ & $0.759\forcese{0.010}$ & $0.37$ & $34.7$ & $11.30$ \\
\midrule
\multicolumn{7}{@{}l}{\textit{Generative world models}} \\
Wan 2.1 I2V 14B & $0.184\forcese{0.019}$ & $0.073\forcese{0.031}$ & $0.756\forcese{0.012}$ & $0.53$ & $34.3$ & $10.71$ \\
DreamZero-DROID & $0.150\forcese{0.016}$ & $0.081\forcese{0.033}$ & $0.720\forcese{0.011}$ & $0.69$ & $33.3$ & $10.33$ \\
Cosmos3-Edge & $0.134\forcese{0.013}$ & $0.043\forcese{0.010}$ & $0.743\forcese{0.010}$ & $0.36$ & $30.5$ & $11.42$ \\
Cosmos3-Edge DROID & $0.132\forcese{0.013}$ & $0.032\forcese{0.008}$ & $0.745\forcese{0.011}$ & $0.34$ & $33.1$ & $11.98$ \\
\midrule
\multicolumn{7}{@{}l}{\textit{Vision--language / action models}} \\
$\pi_0$ & $0.069\forcese{0.009}$ & $-0.014\forcese{0.014}$ & $0.742\forcese{0.017}$ & $0.17$ & $24.3$ & $11.72$ \\
GR00T N1.7 (2 frames) & $0.032\forcese{0.005}$ & $0.020\forcese{0.012}$ & $0.726\forcese{0.017}$ & $0.18$ & $25.2$ & $12.70$ \\
\bottomrule
\end{tabular}}
\end{table}

%% file: tables/joint_selection/table_force_diagnostics.tex
\begin{table}[t]
\centering
\caption{Additional diagnostics on \textbf{Hoi!}. We measure if each models response follows the predicted/target within-episode SD ratio ($A$), Hit rate: \% of peaks within 0.25\,s and MAE$_{90+}$: MAE (N) in the top training-force decile. Sorted by gripper MAE$_{90+}$ within each family. Video models perform well, localizing force peaks successfully in around 50\% of human view cases, with VJEPA2.1 achieving the lowest MAE and best prediction range $A$. DreamZero stands out from the other world models, with the best MAE and a better prediction range $A$ than VJEPA2.1 in the gripper view.}
\vspace{-1mm}
\label{tab:force_diagnostics}
\footnotesize
\setlength{\tabcolsep}{1.8pt}
\begin{tabular}{@{}lcccccc@{}}
\toprule
& \multicolumn{3}{c}{Human view} & \multicolumn{3}{c}{Gripper view} \\
\cmidrule(lr){2-4}\cmidrule(lr){5-7}
Representation & $A\uparrow$ & Hit$\uparrow$ & MAE$_{90+}\downarrow$ & $A\uparrow$ & Hit$\uparrow$ & MAE$_{90+}\downarrow$ \\
\midrule
\multicolumn{7}{@{}l}{\textit{Image and geometry encoders}} \\
DINOv3 ViT-H+ & 0.30 & 31.3 & 8.41 & 0.43 & 35.6 & 6.83 \\
VGGT-Omega 1B & 0.54 & 35.1 & 7.57 & 0.38 & 36.4 & 7.78 \\
\midrule
\multicolumn{7}{@{}l}{\textit{Video encoders}} \\
VideoMAEv2 giant & 0.44 & 47.3 & 7.59 & 0.66 & 37.9 & 6.82 \\
V-JEPA 2.1 ViT-G & 1.02 & 53.4 & 6.25 & 0.72 & 45.5 & 6.84 \\
\midrule
\multicolumn{7}{@{}l}{\textit{Generative world models}} \\
DreamZero-DROID & 0.66 & 39.7 & 6.75 & 0.82 & 42.4 & 6.82 \\
Cosmos3-Nano & 0.34 & 31.3 & 9.36 & 0.49 & 43.9 & 7.46 \\
Cosmos-Predict2 2B & 0.45 & 34.4 & 9.91 & 0.87 & 40.9 & 7.68 \\
Cosmos3-Nano DROID & 0.30 & 31.3 & 9.62 & 0.47 & 35.6 & 7.90 \\
Wan 2.1 1.3B & 0.47 & 32.8 & 8.93 & 0.70 & 34.8 & 8.11 \\
Cosmos3-Edge & 0.24 & 22.9 & 9.70 & 0.41 & 34.8 & 8.37 \\
Cosmos3-Edge DROID & 0.24 & 29.8 & 9.19 & 0.28 & 31.8 & 8.59 \\
Wan 2.1 I2V 14B & 0.32 & 38.9 & 8.54 & 0.47 & 37.1 & 9.30 \\
\midrule
\multicolumn{7}{@{}l}{\textit{Vision--language / action models}} \\
GR00T N1.7 (2 frames) & 0.47 & 35.9 & 7.95 & 0.47 & 31.1 & 8.04 \\
Cosmos-Reason2 (video) & 0.45 & 31.3 & 8.35 & 0.73 & 37.9 & 8.04 \\
OpenVLA-7B & 0.83 & 35.1 & 7.39 & 0.39 & 25.0 & 8.64 \\
$\pi_0$ & 0.15 & 22.1 & 8.12 & 0.52 & 30.3 & 8.71 \\
\midrule
Uniform peak time & -- & 22.4 & -- & -- & 22.7 & -- \\
Constant median & 0 & -- & 10.92 & 0 & -- & 10.91 \\
\bottomrule
\end{tabular}
\vspace{-6mm}
\end{table}

%% file: figures/qualitative/figure_overlay_rh20t.tex
\begin{figure*}[t]
\centering
\includegraphics[width=\textwidth]{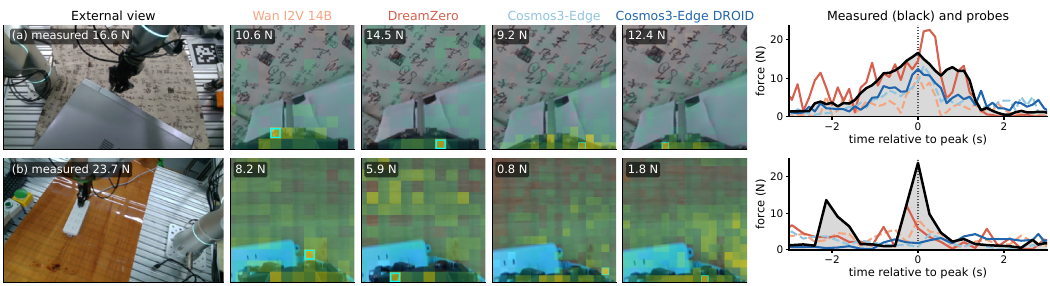}
\caption{Qualitative overlay and evaluation of selected models on two episodes from the test-set of \textbf{RH20T}: (a) opening a microwave door and (b) pressing the button of a power strip. \textit{Left}: external view with GT interaction force. \textit{Middle}: each models input image at the peak time with the affine map before max-pooling overlaid on the image. Brighter tokens correspond to higher scores, with the highlighted patch selected by the max-pooling for the final score. \textit{Right}: force over the recording, with base models dashed and robot-trained models solid. In (a), all four readouts respond at the gripper and the robot-trained models cover more of the force range. In (b), no readout follows the press, and the robot-trained models are no closer than their base models.}
\label{fig:overlay_rh20t}
\vspace{-4mm}
\end{figure*}

%% file: tables/lora_outlook/table_lora_outlook.tex
\begin{table}[t]
\centering
\caption{We fine-tune V-JEPA 2.1 with LoRA adapters on RH20T force labels. Hoi! zero-shot applies the RH20T-trained linear probe unchanged, including its RH20T force normalization. Hoi! probe trains a new readout on each model, recovering force normalization. Training with force as target shows promising results.}
\vspace{-1mm}
\label{tab:lora_outlook}
\footnotesize
\setlength{\tabcolsep}{2.5pt}
\resizebox{\columnwidth}{!}{%
\begin{tabular}{@{}lcccccc@{}}
\toprule
& \multicolumn{2}{c}{RH20T} & \multicolumn{2}{c}{Hoi! zero-shot} & \multicolumn{2}{c}{Hoi! probe} \\
\cmidrule(lr){2-3}\cmidrule(lr){4-5}\cmidrule(lr){6-7}
Encoder & $r\uparrow$ & $R^2\uparrow$ & $r\uparrow$ & $R^2\uparrow$ & $r\uparrow$ & $R^2\uparrow$ \\
\midrule
Frozen (control) & $0.185\forcese{0.018}$ & $0.098\forcese{0.031}$ & $0.033\forcese{0.042}$ & $-0.475\forcese{0.117}$ & $0.427\forcese{0.026}$ & $0.035\forcese{0.104}$ \\
LoRA on RH20T force & $0.325\forcese{0.026}$ & $0.234\forcese{0.048}$ & $0.205\forcese{0.037}$ & $-0.923\forcese{0.564}$ & $0.524\forcese{0.022}$ & $0.098\forcese{0.152}$ \\
\bottomrule
\end{tabular}}
\vspace{-5mm}
\end{table}

%% file: sec/5_conclusion.tex
\section{Conclusion}
\label{sec:conclusion}

We asked whether visual foundation models capture the force involved in an interaction, and probed frozen image and video encoders, video world models, and VLA models on two manipulation datasets with measured force. The answer is largely negative. Most models indicate when contact occurs, but they do not recover how much force is applied. V-JEPA is the exception and the best of the models we probe, yet it explains less than a fifth of the variance in contact force. Robot training does not reliably improve force awareness either: world models and VLA models trained on robot data rarely encode force better than the models they build on. Visual force awareness can be learned, however, as fine-tuning on force labels improves decoding and transfers to a second dataset.

With embodied systems becoming more capable and more widely used, many of them will act around people, requiring knowledge of forces for safe interaction. Force, however, is hidden from view, as are weight and friction, and current world models and policies learn only from what can be observed: appearance, geometry, and motion. Humans learn these hidden variables by interacting with objects and feeling how they respond. We believe that embodied models should follow a similar paradigm, and learn about the physical world from the proprioceptive signals a robot records already during manipulation.

%% file: sec/X_suppl.tex
\clearpage
\setcounter{page}{1}
\maketitlesupplementary
\renewcommand{\topfraction}{0.9}
\renewcommand{\dbltopfraction}{0.9}
\renewcommand{\textfraction}{0.07}
\renewcommand{\floatpagefraction}{0.75}
\renewcommand{\dblfloatpagefraction}{0.75}
\setcounter{topnumber}{3}
\setcounter{dbltopnumber}{3}
\setcounter{totalnumber}{5}

This supplement is organized as follows. Sec.~\ref{sec:supp_data} details the datasets and the evaluation protocol, Sec.~\ref{sec:supp_models} the feature extraction for each model family, and Sec.~\ref{sec:supp_repro} the training setup and compute. Sec.~\ref{sec:supp_results} reports additional results: contact detection on Hoi!, test-only scores, peak and force-range diagnostics on Hoi!\ and RH20T, and zero-shot estimates from multimodal LLMs. Sec.~\ref{sec:supp_robustness} tests whether the probe design explains the low force decoding, Sec.~\ref{sec:supp_lora} describes the force fine-tuning of V-JEPA, Sec.~\ref{sec:supp_qualitative} shows readout maps for all models, and Sec.~\ref{sec:supp_depth} traces force information across model depth.

\section{Datasets and evaluation protocol}
\label{sec:supp_data}

\paragraph{Hoi!.}
The six evaluation environments of the Hoi!\ benchmark are divided into validation (bedroom\_6, livingroom\_1, office\_1) and test (bathroom\_2, bedroom\_4, kitchen\_7); all other locations are used for training. The division into validation and test is ours. Validation contains four recordings and test contains five, so the pooled evaluation covers six scenes and nine recordings per view. Each window covers 64 frames (8.5\,s, Sec.~4.1 of the main paper). For training, a new window starts every 32 frames, so consecutive training windows overlap by half. For evaluation, a new window starts every 64 frames, so the windows tile each recording without overlap and every time step is evaluated once. Only the last window of a recording is shifted back to end with the recording, and where it overlaps the previous window, the two predictions are averaged. The window predictions are joined into one trajectory per recording on a common 4\,Hz reference, and all metrics are computed on these trajectories. RH20T uses the same windows. Models that run at 16\,Hz are mapped to the reference by the nearest time step, without interpolation; the largest offset is 42\,ms. VLA models receive the second image of each pair of frames, with the force target of the pair unchanged. Table~\ref{tab:dataset_stats} summarizes both datasets.

\input{tables/datasets/table_dataset_stats}

\paragraph{RH20T and the task prior.}
The RH20T split holds out operator--task groups, so 105 of the 109 test tasks also occur in training. A model that only recognizes the task could therefore output the typical force of that task without observing the contact. The task prior in Table~2 of the main paper measures this shortcut. It receives the task identity but no video and predicts the average force of that task in the training recordings, in three variants: constant in time, as a function of the time since the start of the recording (bins of 0.5\,s), and as a function of the elapsed fraction of the recording (50 bins). The last variant uses the length of the recording, which no probed model observes. Tasks that do not occur in training use the average over all training recordings. As for the probes, each metric is reported for the variant selected on validation, and the prior is scored on the same windows, targets, and contact episodes. It reaches $r=0.02$ and a negative contact $R^2$, so the probed models do not owe their scores to task recognition.

\paragraph{Contact episodes.}
We threshold the force trajectory at 2\,N, merge gaps of at most four time steps, discard segments shorter than two steps, and pad each segment by two steps on both sides. Padding and merging keep the rise and release of force inside an episode, which the peak and amplitude diagnostics need, and prevent a brief dip below 2\,N from splitting one interaction into several episodes. Episodes shorter than four steps are excluded from the mean correlation; magnitude metrics use all episodes. The pooled Hoi!\ evaluation contains 131 human-view and 132 gripper-view episodes with about 2,090 time steps each. The segmentation only defines evaluation boundaries and is never an input to a model or probe. Sec.~\ref{sec:supp_strict} evaluates the time steps above 2\,N alone.

\paragraph{Bootstrap.}
Standard errors are computed from 10,000 bootstrap draws (seed 17) that resample whole clusters with replacement: scenes on Hoi!\ and operator--task groups on RH20T. Each drawn cluster keeps all its episodes, and the same draws are shared by all models. Correlation is the mean over the drawn episodes, while contact $R^2$ and MAE are recomputed from the drawn time steps. The errors describe the variability of the evaluation data for fixed probes and fixed selections and do not include probe training or selection. On Hoi!, the draws resample the six evaluation scenes. Table~\ref{tab:bootstrap} compares this with resampling the nine recordings. Both give similar errors, so the choice of cluster does not affect the comparisons in the main paper. On RH20T, the draws resample the 247 operator--task groups of the evaluation splits (25 validation, 222 test) with 2,657 recordings and 6,600 contact episodes. With about 40 times more clusters than on Hoi!, the RH20T errors are considerably smaller, and the standard error of contact $R^2$ is two to seven times smaller for every model (Table~2 of the main paper).

\FloatBarrier
\section{Models and feature extraction}
\label{sec:supp_models}

\paragraph{Image and video encoders.}
Video encoders are probed on their normalized patch tokens after every block, VGGT-$\Omega$ on the concatenated frame and global aggregator outputs. The supplement-only checkpoints V-JEPA~2 ViT-L, V-JEPA~2.1 ViT-L, VideoMAEv2 ViT-L, and LeVJEPA ViT-L use the same protocol as their larger counterparts. LeVJEPA predicts on a finer temporal grid, and we average its predictions over adjacent pairs to match the common reference.

To test whether explicit frame-to-frame change helps a per-frame image encoder, we also probe DINOv3 on the mean of two adjacent frame features concatenated with their difference,
\begin{equation}
 \mathbf{z}^{\mathrm{diff}}_{t,k}
 = \left[\frac{\mathbf{h}_{2t,k}+\mathbf{h}_{2t+1,k}}{2}\,;\,
 \mathbf{h}_{2t+1,k}-\mathbf{h}_{2t,k}\right].
\end{equation}
This control (DINOv3 +diff) uses the same readout, layer sweep, and selection as DINOv3 and does not improve decoding: pooled $r$ is $0.155$ and $0.321$ in the human and gripper views, compared with $0.173$ and $0.406$ for DINOv3.

\paragraph{Video world models.}
World models are probed on their video-latent tokens. Cosmos3-Edge receives 512-pixel inputs to obtain a $16\times16$ token grid, whereas Wan~2.1~1.3B and Cosmos-Predict2 obtain this grid from 256-pixel inputs. Wan I2V~14B runs at $352\times640$ with first-frame conditioning and full attention over the video. Its autoencoder yields 33 latent steps, of which the first serves only as context, and we probe blocks 8, 16, 24, 32, and 40. Cosmos3-Nano and Nano DROID attend over all 33 latents and the text stream; we probe blocks 7, 14, 22, 29, and 36. DreamZero is probed on its normalized visual-context tokens ($22\times40$ grid) at blocks 8, 16, 24, 32, and 40. Its attention is causal between chunks of two latents, with a rolling cache of nine latents, and each Hoi!\ camera fills the input canvas on its own, unlike the three-camera layout used for DROID training. The probed states precede DreamZero's action prediction, so we probe its visual representation and not a complete policy. Because DreamZero processes video causally while Wan attends over the full window, the Wan/DreamZero comparison does not isolate robot training.

\paragraph{VLA models.}
In VLA models, we probe the image tokens before and after the projection into the language model, after every language-model block, and after the final normalization. GR00T~N1.7 additionally has a policy adapter, which we probe as well; it precedes the fusion with robot state and the action decoder. GR00T receives two images one second apart from the same camera, as specified by its DROID checkpoint, and no robot state. Cosmos-Reason2 receives the window as video through its released video processor, which merges pairs of frames, and generates no text. For a matched comparison with GR00T, which keeps 16 of the 28 language-model blocks of Cosmos-Reason2, we also probe the original Cosmos-Reason2-2B weights truncated to the same 16 blocks with the same image and text processing. Both models are additionally probed with a single image per time step. Table~\ref{tab:bootstrap} lists the correlation of every variant, and Figs.~\ref{fig:policy_other_stages}--\ref{fig:policy_reason2_full_image} compare them across extraction stages.

\FloatBarrier
\section{Reproducibility}
\label{sec:supp_repro}
Hyperparameters and input context are matched within but not across model families, as each model is run with its own released inference protocol (Sec.~\ref{sec:supp_models}). The only deviation within a pair is the training batch of the Cosmos3-Edge DROID probe (32 instead of 8). Every run uses a single GPU. Feature extraction, probe training, and LoRA fine-tuning run on NVIDIA H200 GPUs (141\,GB); some smaller image and video encoders on Hoi!\ and the open-weight multimodal LLMs run on an NVIDIA RTX~3090 (24\,GB).

\FloatBarrier
\section{Additional results}
\label{sec:supp_results}

\subsection{Contact detection}
\label{sec:supp_auroc}
\input{tables/contact_detection/table_contact_detection_hoi}
Table~\ref{tab:contact_detection} reports contact detection on Hoi!\ for every model. All probes detect contact well above chance, and the confidence intervals of all models exclude $0.5$. In the gripper view, the single-image encoders VGGT-$\Omega$ ($0.92$) and DINOv3 ($0.90$) reach the highest AUROC, on par with V-JEPA ($0.89$) and Cosmos-Reason2 ($0.88$). In the human view, where the contact region is small and often occluded by the hand, the world models fall behind: Cosmos3 and Wan~2.1~1.3B stay between $0.60$ and $0.66$, while most other models reach $0.80$ or more. Detecting contact is only loosely related to tracking force. Across the 16 models of the main tables, the Spearman correlation between AUROC and $r$ is $0.32$ in the gripper view and $0.49$ in the human view, and V-JEPA, which tracks force best, has one of the lowest human-view AUROCs ($0.74$).

\subsection{Test-only results and bootstrap}
\label{sec:supp_test}
\input{tables/joint_selection/table_test_views}
\input{tables/joint_selection/table_bootstrap_views}
The main tables pool validation and test scenes, which doubles the number of evaluated episodes but reuses the validation scenes for selection. Table~\ref{tab:force_test} reports the test scenes alone (65 episodes per view). The ordering is unchanged: in the gripper view, the V-JEPA checkpoints track force best and are the only ones with positive contact $R^2$, followed by the other video encoders and DreamZero. RH20T does not depend on this choice. Its test split holds 6,141 of the 6,600 pooled episodes, so test-only and pooled correlations differ by at most 0.005 (Sec.~\ref{sec:supp_rh20t}), and the conclusions of the main paper hold on this much larger evaluation. Table~\ref{tab:bootstrap} gives the bootstrap errors of the pooled Hoi!\ correlation together with percentile intervals.

\subsection{Peak and force-range diagnostics}
\label{sec:peak_diagnostics}
\input{tables/joint_selection/table_peaks_views}
\input{tables/joint_selection/table_range_bins_views}
Tables~\ref{tab:peak_diagnostics} and~\ref{tab:range_bins} extend the diagnostics of Table~3 in the main paper. The peak tables compare the predicted and measured maximum of each contact episode. For episode $E$, the timing offset is
\begin{equation}
 \Delta t_E=0.25\left(\arg\max_{t\in E}\hat y_t-\arg\max_{t\in E}y_t\right)\ \mathrm{s},
\end{equation}
and the hit rates count episodes with $|\Delta t_E|$ of at most one or two time steps. A peak time drawn uniformly within each episode serves as timing reference, and the median training peak force as magnitude reference. The range tables report the amplitude ratio $A$, the largest prediction, and the MAE in bins of the training force distribution (percentiles 50, 75, and 90, computed separately for each view).

\paragraph{Peak timing.}
Every model locates the force peak more often than the uniform reference ($22\%$ within one step), but no model does so reliably. The video encoders are best, with V-JEPA~2.1 ViT-G at $53\%$ in the human view and the V-JEPA~2.1 ViT-L at $51\%$ in the gripper view. Most world models and VLA models stay between $30\%$ and $40\%$, and the median timing error of most models is 0.5\,s, two time steps.

\paragraph{Peak magnitude.}
All models estimate the episode peak more accurately than the constant median peak in MAE (4.0--6.5\,N vs.\ 6.7\,N), but not in RMSE (6.3--9.2\,N vs.\ 7.9\,N). The probes thus get typical peaks roughly right and miss the large ones, in line with the compressed amplitude reported in the main paper.

\paragraph{Error across the force range.}
The force bins show where the errors come from. Below the training median, the constant prediction is more accurate than every probe ($0.61$\,N vs.\ at least $0.62$\,N), because the probes respond to the approach of the gripper and to motion without contact. In the top decile, every probe is more accurate than the constant ($5.9$--$9.9$\,N vs.\ $10.9$\,N), but the best ones still miss by about 6\,N. The largest predictions also separate the models. V-JEPA~2.1 ViT-G predicts up to 28\,N in the human view, and DreamZero and Cosmos-Predict2 exceed 12\,N in the gripper view 37 and 112 times. Cosmos-Predict2 nevertheless has a larger top-decile error than DreamZero ($7.7$ vs.\ $6.8$\,N), so its large predictions often occur at the wrong time. A large amplitude alone therefore does not indicate that a model recovers force magnitude.

\subsection{Diagnostics on RH20T}
\label{sec:supp_rh20t}
\input{tables/rh20t/table_rh20t_r_by_peak}
\input{tables/rh20t/table_rh20t_peaks}
Tables~\ref{tab:rh20t_r_by_peak} and~\ref{tab:rh20t_peaks} repeat the Hoi!\ diagnostics on RH20T. The test split holds 6,141 of the 6,600 pooled episodes, so test-only and pooled correlations differ by at most 0.005. The operator--task intervals of all video encoders and world models lie above those of the image encoders and VLA models.

\paragraph{Correlation and contact strength.}
Nearly half of the RH20T episodes peak below 5\,N, and on these weak contacts no model exceeds $r=0.04$. Correlation rises with the peak force for the video encoders and world models, up to $r=0.48$ for V-JEPA and Wan I2V above 20\,N, while the image encoders and VLA models stay at or below $0.16$. The low overall correlations on RH20T thus come from the weak contacts, and the differences between families come from the forceful ones.

\paragraph{Peaks and force range.}
As on Hoi!, the video encoders and Wan I2V locate the peak most often ($34$--$36\%$ within one step, against $25\%$ for a uniform guess), with a median offset of two steps. Peak magnitude is harder than on Hoi!, because RH20T forces have a long tail: the median training peak is 5.3\,N, but measured forces reach 557\,N. Every probe beats the constant median peak in MAE only slightly ($9.7$--$11.3$ vs.\ $11.3$\,N), and none in MedAE ($2.9$--$4.4$ vs.\ $2.8$\,N). Below the training median, the constant force is again more accurate than every probe ($0.53$ vs.\ at least $1.55$\,N). In the top decile, every probe is more accurate than the constant ($9.6$--$12.7$ vs.\ $16.5$\,N), but no prediction exceeds 56\,N.

\subsection{Zero-shot estimates from multimodal LLMs}
\label{sec:supp_vlm}
\input{figures/vlm/figure_vlm_peak_scatter}
Our probes are trained on force labels. A natural question is whether large multimodal models can estimate force without any training, from the video alone. We asked four such models, GPT-6 Astra, Claude Fable 5.1, Gemini 3.1 Pro, and Qwen3.5-9B, for the peak force of each of the 131 human-view contact episodes of the main evaluation. Each model receives 16 frames sampled in order from the episode, at up to 448$\times$448 pixels, together with a short description of the setup: a first-person video of a person manipulating an object with a handheld gripper whose force sensor records the contact force. The prompt asks for the peak contact force in newtons as a single number. The models receive no examples and no force scale, and all run with their reasoning mode enabled. Because they give one number per episode, we compare them to the measured episode peak, \ie the maximum of the per-step force target of the probes within the episode, and not to the force trajectory.

None of the models estimates force magnitude reliably (Fig.~\ref{fig:vlm_peak}). GPT-6 Astra orders the episodes best ($r=0.46$ across episodes) but overestimates almost every peak, with an MAE of 11.2\,N. Claude Fable 5.1 has the lowest error (5.7\,N) but carries almost no information about which episode is forceful ($r=0.09$). Gemini 3.1 Pro answers 0\,N for 77 of the 131 episodes, and Qwen3.5-9B gives a parseable answer for only 94 episodes, 75 of which are either 5 or 10\,N. For comparison, predicting the median training peak for every episode gives an MAE of 6.7\,N. Without supervision, these models fall back on a plausible range of values rather than on what the video shows. This agrees with QuantiPhy (cited in the main paper), which finds that vision--language models give plausible but inaccurate numerical estimates of physical quantities.

\section{Robustness of the probe}
\label{sec:supp_robustness}
The main paper finds that force magnitude is poorly decodable from every model. Here we test whether the evaluation or the probe causes this.

\subsection{Strict-contact evaluation}
\label{sec:supp_strict}
\input{tables/strict_contact/table_strict_contact}
The padded contact episodes contain time steps below 2\,N, so a probe that only responds to the onset and release of contact obtains a positive within-episode correlation. Table~\ref{tab:strict_contact} therefore repeats the evaluation on only the time steps at or above 2\,N, with unchanged probes. On RH20T and in the Hoi!\ human view, correlation is almost unchanged. In the Hoi!\ gripper view, it roughly halves for all models, and DINOv3 loses the most ($0.41$ to $0.13$). There, DreamZero has the highest correlation ($0.33$), followed by V-JEPA ($0.27$) and VideoMAEv2 ($0.26$). Contact $R^2$ decreases for every model. V-JEPA remains the only model with positive $R^2$ on Hoi!\ ($0.05$ in the gripper view) and has the highest $R^2$ on RH20T ($0.11$). The single-image encoders thus owe most of their Hoi!\ gripper-view correlation to contact onset and release, while the video models also follow force during contact.

\subsection{Calibration and training objective}
\label{sec:supp_calibration}
\input{tables/strict_contact/table_calibration_bound}
\input{tables/strict_contact/table_loss_objective}
The probes are trained on complete trajectories, most of which contain no contact, while contact $R^2$ and the amplitude are evaluated inside contact episodes. We test in two ways whether this objective, and not the representation, causes the low contact $R^2$ and the compressed force range.

\paragraph{Calibration bound.}
A probe could track force correctly but report it at the wrong scale or offset. We therefore fit a slope $a$ and an offset $b$ by least squares so that $a\,\widehat{y}_t+b$ matches the measured force on the contact time steps, and recompute contact $R^2$. Because the fit uses the evaluated time steps themselves, the result is the highest contact $R^2$ that any rescaling of the probe output can reach. Table~\ref{tab:calibration_bound} shows that calibration does not change the conclusion. On Hoi!, no calibrated probe exceeds $R^2=0.06$ except V-JEPA in the gripper view ($0.19$), and on RH20T no frozen model exceeds $0.14$. On Hoi!, all slopes are below one, so the probe outputs vary more than their information supports. On RH20T, the slopes are close to one. A predictor with little information about the target has to predict close to the mean, so the compressed force range follows from the weak force information in the features. Calibration does narrow the differences between models on RH20T: the other models underestimate contact force by a constant offset, and once this is corrected, DreamZero ($0.13$), Wan I2V~14B ($0.13$), and VideoMAEv2 ($0.12$) are close to V-JEPA ($0.14$). Part of V-JEPA's higher $R^2$ on RH20T thus reflects better calibration and not only better tracking of force magnitude.

\paragraph{Alternative objectives.}
Calibration rescales the probe output but cannot change which feature direction the probe reads. We therefore retrain the readout on the final-layer features of V-JEPA (both views) and of VideoMAEv2, DINOv3, and VGGT-$\Omega$ (human view) with three alternative objectives. The first keeps all time steps and replaces the smooth-$L_1$ loss with a squared error. The other two compute the smooth-$L_1$ loss or the squared error only on contact time steps, \ie those with a measured force of at least 2\,N, so that the probe is no longer trained to predict zero force outside contact. Targets, optimizer, sampling, and the number of epochs are those of the main probe, and every readout, including ours, uses the epoch with the best validation contact $R^2$. Table~\ref{tab:loss_objective} shows that no alternative improves contact $R^2$: our objective gives the highest $R^2$ for every model and view, and the highest correlation in four of five settings. The alternatives increase the amplitude, up to $A=1.19$, but the larger responses are less accurate. The compressed force range under our objective therefore does not hide force magnitude that a different loss would recover.

\subsection{Location selection}
\label{sec:supp_selection}
\input{tables/experiment_findings/figure_location_selection}
\input{tables/experiment_findings/table_selection_contact_r2}
We select a separate location for correlation ($\ell^*_r$) and for contact $R^2$ ($\ell^*$), both on validation (Sec.~3.4 of the main paper). Selecting a single location by validation $R^2$ for all metrics would understate tracking, because the location that best fits force values on complete trajectories is not always the one that best follows force within episodes. Fig.~\ref{fig:location_selection} compares the two rules. Correlation-based selection changes 43 of the selected locations and improves test correlation in 33 of the model/view comparisons, lowers it in ten, and leaves five unchanged. Table~\ref{tab:selection_contact_r2} shows that contact $R^2$ at the correlation-selected location is lower for most models, which is why we report it at $\ell^*$. Neither rule uses test data, and the conclusions of the main paper hold under both.

\subsection{Attentive readouts and training seeds}
\label{sec:attentive_robustness}
\input{tables/attentive_robustness/table_attentive_robustness}
As is standard for frozen representations, our probe is linear, so that differences between models reflect the representations and not the readout. To check that a nonlinear readout does not change the results, we train an attentive probe on the final-layer human-view features of the image and video encoders. The probe projects tokens to 256 dimensions, applies two blocks of factorized spatial and temporal attention, and pools with a learned query, with about 4.3 million trainable parameters. We trained it with three seeds and 36 configurations, varying width, temporal context, pooling, and attention factorization. The configuration was chosen on V-JEPA and shared across models. Training uses the same data, loss, and sampling as the main probe, with learning rate $10^{-4}$ and weight decay $10^{-4}$.

Table~\ref{tab:attentive_robustness} compares the attentive probe with the affine probe on the same final-layer features. For V-JEPA, the attentive probe raises correlation from $0.37$ to $0.44\pm0.02$ and contact $R^2$ from $-0.14$ to $0.07\pm0.04$, and all three seeds exceed every other model. For the other models, contact $R^2$ stays negative, and VideoMAEv2 and VGGT-$\Omega$ even lose correlation. More probe capacity or context is not uniformly beneficial: with one, two, and 32 tubelets of temporal context, V-JEPA reaches $0.42$, $0.44$, and $0.41$. A richer readout thus extracts somewhat more from V-JEPA but does not change the ranking or the overall picture of weak force decoding.

\section{Force fine-tuning}
\label{sec:supp_lora}
For Sec.~4.5 of the main paper, we add LoRA adapters (rank 16, $\alpha=32$) to the attention (qkv and output projection) and MLP layers of every block of V-JEPA~2.1 ViT-G, 15.1M trainable parameters in total. The adapters start as an identity, so training begins from the frozen encoder. The readout is the probe of the main paper on the final block, and data, loss, sampling, and target normalization are those of the frozen RH20T probes. We train for five epochs with batch size 24, a constant learning rate of $10^{-3}$ for the readout, and warmup followed by cosine decay for the adapters. The adapter learning rate was chosen among $3\cdot10^{-5}$, $10^{-4}$, and $3\cdot10^{-4}$ by validation $R^2$ after the first epoch, which selected $3\cdot10^{-4}$. We evaluate twice per epoch and keep the adapters with the best validation $R^2$. The frozen control in Table~4 of the main paper is the frozen probe on the same final block.

For \emph{Hoi!\ zero-shot}, we apply the RH20T-trained readout and its RH20T force normalization to Hoi!\ without any change. For \emph{Hoi!\ probe}, we train a new readout on the adapted and on the frozen encoder with the Hoi!\ protocol of the main paper.

\FloatBarrier
\section{Qualitative results}
\label{sec:supp_qualitative}
\begingroup\input{figures/qualitative/supp/figures_supp_overlays}\endgroup
Figures~\ref{fig:supp_overlay_hoi} and~\ref{fig:supp_overlay_rh20t} show where the readout of every model responds, at the force peak of the recordings in Figs.~3 and~4 of the main paper. For Hoi!, we show the human and the gripper view at the same instant, which share the force measurement. Each tile uses the probe and location of the main tables, and the readout in the tile equals the prediction that enters the reported metrics.

\paragraph{Where the readout responds.}
The image and video encoders and most world models place their highest score on the gripper or the contact region, in both datasets and both Hoi!\ views. VGGT-$\Omega$ and the VLA models respond across large parts of the scene, and the token they select lies less consistently at the contact, in line with their weak force tracking in the main tables. A readout at the contact is necessary but not sufficient: most models that look at the gripper still predict a fraction of the measured peak.

\paragraph{The view matters.}
The same contact can be read very differently from the two Hoi!\ views. During the 20.9\,N push in Fig.~\ref{fig:supp_overlay_hoi} (bottom), V-JEPA predicts 18.2\,N from the human view but 6.4\,N from the gripper view, while during the 17.0\,N contact (top) the gripper view gives the higher readout (9.0 vs.\ 4.5\,N). Which view shows the force best thus depends on the interaction, and averaged over episodes, the gripper view is easier for every model (Table~1 of the main paper).

\paragraph{RH20T.}
On the microwave door (Fig.~\ref{fig:supp_overlay_rh20t}, top), V-JEPA overshoots the measured peak (20.5 vs.\ 16.6\,N), and VideoMAEv2 and DreamZero recover about two thirds of it, while DINOv3, VGGT-$\Omega$, and the VLA models stay below 7\,N. On the power strip (bottom), where no world model follows the press (Fig.~4 of the main paper), V-JEPA is the only model that recovers more than half of the peak (13.2 vs.\ 23.7\,N). These are single examples, but they agree with the ranking of the main tables.

\paragraph{Supplementary video.}
The accompanying video shows two Hoi!\ contacts of Fig.~3 (b and c) and the two RH20T recordings of Fig.~4 of the main paper over time. Each clip shows both cameras (the human and gripper views on Hoi!, the external and in-hand cameras on RH20T), the readout map of each model on its input frames, and the measured and predicted force over the recording. Maps are shown only around contact and are normalized per frame, and a box marks the token selected by max pooling. Each clip plays first at normal speed and then at half speed, first for the world-model pairs (Wan~I2V~14B and DreamZero, Cosmos3-Edge and Cosmos3-Edge DROID) and then for V-JEPA~2.1 and VideoMAEv2. The Hoi!\ office contact (human view, 20.9\,N peak) is a mixed case: V-JEPA (18.2\,N) and DreamZero (13.1\,N) follow the force, while both Cosmos3-Edge models stay below 4\,N. The sustained push in the bathroom (gripper view, 19.6\,N) is a failure case, in which no model predicts more than 7\,N. On RH20T, the microwave door is a success case, in which all readouts respond at the gripper. The power-strip press is a failure case for the world models: none of them follows the press, whereas V-JEPA does and VideoMAEv2 barely responds.

\FloatBarrier
\section{Force information across depth}
\label{sec:supp_depth}
Figures~\ref{fig:encoders_test_r}--\ref{fig:world_conditions} show test correlation and contact $R^2$ at every probed location. Each point is a single probe trained at that location, without smoothing and without uncertainty, so neighboring blocks can differ by $0.1$ or more. Stars mark the locations selected on validation. We discuss the gripper view unless noted; the human view follows the same trends at a lower level.

\paragraph{Image and video encoders.}
The video encoders build up force information with depth (Fig.~\ref{fig:encoders_test_r}). V-JEPA~2.1 ViT-G rises from $r=0.13$ after the first block to about $0.5$ between blocks 23 and 35. It drops to $0.23$ in block 39 and recovers to $0.42$ at the output, so the last blocks before the output carry less force information than the middle of the network. VideoMAEv2 giant rises from $0.19$ to about $0.4$ by block 13 and then stays level. The smaller V-JEPA checkpoints show the same rise, with V-JEPA~2.1 ViT-L reaching its maximum in block 22 of 24. DINOv3 also rises with depth, from $0.07$ to about $0.35$, but its profile is noisier. VGGT-$\Omega$ is the exception: it starts at $0.27$ and declines in its later blocks, which are trained for geometry. In the human view, V-JEPA only reaches its level in the last few blocks (block 34: $0.36$), whereas VideoMAEv2 is already level from block 16. Contact $R^2$ follows correlation (Fig.~\ref{fig:encoders_test_r2}) and is positive only in the later blocks of the V-JEPA checkpoints.

\paragraph{Video world models.}
The world models do not build up force information (Figs.~\ref{fig:world_wan_dreamzero}--\ref{fig:world_cosmos_predict2}). Wan~2.1~1.3B is best after the first block ($0.41$), directly behind the video autoencoder, and stays between $0.2$ and $0.35$ afterwards. Cosmos3-Edge peaks in block 5 ($0.34$) and falls to about $0.2$ in the second half of the network. Cosmos-Predict2 fluctuates between $0.14$ and $0.37$ without a trend. The diffusion transformers thus add little to what their autoencoder latents already provide. The human view makes this clearer: Cosmos3-Edge stays near $r=0$ at every block. DreamZero and Wan I2V~14B, probed at five depths, differ at the output: DreamZero is best at its last block ($0.45$), where Wan I2V~14B is worst ($0.16$). DreamZero's action prediction reads from these late states, and we see this as consistent with our hypothesis that joint action prediction keeps force-related information in the video features, although five depths of a single probe do not establish it.

\paragraph{Prompt and noise conditions.}
Fig.~\ref{fig:world_conditions} repeats the sweeps of Wan~2.1~1.3B, Cosmos-Predict2, and Cosmos3-Edge under six further conditions. Three change the text prompt and keep the input clean: a static prompt (``Manipulate the drawer.'', the instruction the VLA models receive), the labelled action of each window (``Open the drawer.'', ``Close the drawer.'', ``Open the door.'', or ``Close the door.''), and a swapped action with the opposite verb and the same object. The action prompt uses annotations that the video alone does not provide, so it is an oracle, and the swapped prompt is its control. The other three keep the empty prompt and add noise to the clean video latent $\mathbf{x}_0$ as $(1-t/1000)\,\mathbf{x}_0+(t/1000)\,\boldsymbol{\epsilon}$ with Gaussian noise $\boldsymbol{\epsilon}$ and $t=50$, $100$, or $200$. The model receives the matching timestep, the noise is fixed for each window, and, as in all conditions, the model runs a single forward pass without denoising. Text has no consistent effect on any model. Noise has none on Wan~2.1~1.3B and Cosmos-Predict2, but it raises the correlation of Cosmos3-Edge in its later blocks from about $0.2$ to $0.35$--$0.45$. We did not run this condition for the other models and report it as an observation. The main tables use the clean, empty-prompt condition for all world models, following prior probing work.

\paragraph{VLA models.}
In VLA models, force information is weak where the image leaves the vision encoder (Figs.~\ref{fig:policy_other_stages}--\ref{fig:policy_reason2_full_image}). Before the projection into the language model, gripper-view correlation is $0.10$ for OpenVLA, $0.15$ for $\pi_0$, $0.18$ for GR00T, and $0.10$--$0.15$ for Cosmos-Reason2, well below the later blocks of DINOv3. The language model raises it moderately in its first half: Cosmos-Reason2 reaches $0.34$ in block 10 with image input and $0.35$ in block 8 with video input. The second half loses this again, and with video input, correlation is close to zero from block 23 on. $\pi_0$ stays between $0.05$ and $0.23$ and drops to $-0.09$ in its last block, and GR00T stays between $0.13$ and $0.29$ at all stages, including its policy adapter. The language model therefore does not recover force information that the vision encoder lacks, and its later blocks, which prepare text or action outputs, discard part of what is there.

\clearpage
\input{figures/probe_locations/supplementary_figures}

%% file: tables/datasets/table_dataset_stats.tex
\begin{table}[tp]
\centering
\footnotesize
\setlength{\tabcolsep}{3pt}
\caption{Statistics of the \textbf{Hoi!} and \textbf{RH20T} splits. Windows span 64 frames (8.5\,s). Groups are recording locations (Hoi!) or operator--task pairs (RH20T). Hoi!\ contact episodes are given for the human\,/\,gripper view.}
\label{tab:dataset_stats}
\begin{tabular}{@{}llrrrrr@{}}
\toprule
Dataset & Split & Windows & Recordings & Groups & Hours & Contact ep. \\
\midrule
Hoi! & train & 724 & 13 & 12 & 0.83 & 275 / 274 \\
 & val & 109 & 4 & 3 & 0.24 & 66 / 67 \\
 & test & 151 & 5 & 3 & 0.33 & 65 / 65 \\
\midrule
RH20T & train & 7,241 & 908 & 438 & 13.0 & 2,350 \\
 & val & 1,093 & 218 & 25 & 3.8 & 459 \\
 & test & 11,768 & 2,439 & 222 & 35.3 & 6,141 \\
\bottomrule
\end{tabular}
\end{table}

%% file: tables/contact_detection/table_contact_detection_hoi.tex
\begin{table}[tp]
\centering
\caption{Contact detection on \textbf{Hoi!} with scene-cluster bootstrap standard errors. We report the AUROC of the predicted force for separating contact time steps ($\geq$2\,N) from all others, at $\ell^*_r$. Chance is $0.5$. All models detect contact well above chance.}
\label{tab:contact_detection}
\footnotesize
\setlength{\tabcolsep}{4pt}
\begin{tabular}{@{}lcc@{}}
\toprule
Representation & Human view & Gripper view \\
\midrule
\multicolumn{3}{@{}l}{\textit{Image and geometry encoders}} \\
DINOv3 ViT-H+ & $0.835\forcese{0.026}$ & $0.901\forcese{0.020}$ \\
VGGT-Omega 1B & $0.816\forcese{0.027}$ & $0.916\forcese{0.018}$ \\
\midrule
\multicolumn{3}{@{}l}{\textit{Video encoders}} \\
V-JEPA 2.1 ViT-G & $0.741\forcese{0.028}$ & $0.887\forcese{0.023}$ \\
VideoMAEv2 giant & $0.839\forcese{0.030}$ & $0.877\forcese{0.022}$ \\
\midrule
\multicolumn{3}{@{}l}{\textit{Generative world models}} \\
DreamZero-DROID & $0.806\forcese{0.028}$ & $0.841\forcese{0.025}$ \\
Cosmos3-Nano & $0.608\forcese{0.039}$ & $0.829\forcese{0.026}$ \\
Cosmos3-Nano DROID & $0.601\forcese{0.038}$ & $0.787\forcese{0.036}$ \\
Cosmos-Predict2 2B & $0.655\forcese{0.031}$ & $0.733\forcese{0.032}$ \\
Cosmos3-Edge & $0.613\forcese{0.043}$ & $0.774\forcese{0.022}$ \\
Wan 2.1 I2V 14B & $0.764\forcese{0.026}$ & $0.803\forcese{0.030}$ \\
Cosmos3-Edge DROID & $0.651\forcese{0.034}$ & $0.841\forcese{0.017}$ \\
Wan 2.1 1.3B & $0.637\forcese{0.041}$ & $0.724\forcese{0.040}$ \\
\midrule
\multicolumn{3}{@{}l}{\textit{Vision--language / action models}} \\
Cosmos-Reason2 (video) & $0.828\forcese{0.026}$ & $0.875\forcese{0.022}$ \\
GR00T N1.7 (2 frames) & $0.836\forcese{0.025}$ & $0.845\forcese{0.032}$ \\
$\pi_0$ & $0.823\forcese{0.045}$ & $0.861\forcese{0.024}$ \\
OpenVLA-7B & $0.708\forcese{0.029}$ & $0.797\forcese{0.012}$ \\
\midrule
\multicolumn{3}{@{}l}{\textit{Supplement-only checkpoints}} \\
V-JEPA 2.1 ViT-L & $0.813\forcese{0.021}$ & $0.896\forcese{0.017}$ \\
V-JEPA 2 ViT-L & $0.770\forcese{0.031}$ & $0.883\forcese{0.017}$ \\
LeVJEPA ViT-L & $0.817\forcese{0.033}$ & $0.836\forcese{0.025}$ \\
VideoMAEv2 ViT-L & $0.761\forcese{0.031}$ & $0.864\forcese{0.029}$ \\
\bottomrule
\end{tabular}
\end{table}

%% file: tables/joint_selection/table_test_views.tex
\begin{table}[tp]
\centering
\caption{Force decoding on the \textbf{Hoi!} test scenes only (65 contact episodes per view). We report $r$ at $\ell^*_r$ and contact $R^2$ and MAE at $\ell^*$.}
\label{tab:force_test}
\footnotesize
\setlength{\tabcolsep}{2.5pt}
\resizebox{\columnwidth}{!}{%
\begin{tabular}{@{}lrrrrrr@{}}
\toprule
& \multicolumn{3}{c}{Human view} & \multicolumn{3}{c}{Gripper view} \\
\cmidrule(lr){2-4}\cmidrule(lr){5-7}
Representation & $r\uparrow$ & $R^2\uparrow$ & MAE$\downarrow$ & $r\uparrow$ & $R^2\uparrow$ & MAE$\downarrow$ \\
\midrule
\multicolumn{7}{@{}l}{\textit{Image and geometry encoders}} \\
DINOv3 ViT-H+ & 0.119 & $-$0.048 & 2.799 & 0.366 & $-$0.037 & 2.914 \\
VGGT-Omega 1B & 0.211 & $-$0.054 & 2.948 & 0.372 & $-$0.028 & 2.873 \\
DINOv3 +diff & 0.115 & $-$0.105 & 2.915 & 0.227 & $-$0.044 & 2.874 \\
\midrule
\multicolumn{7}{@{}l}{\textit{Video encoders}} \\
V-JEPA 2.1 ViT-G & 0.249 & $-$0.112 & 3.121 & 0.500 & 0.119 & 2.804 \\
V-JEPA 2 ViT-L & 0.259 & $-$0.134 & 3.201 & 0.471 & 0.063 & 2.861 \\
V-JEPA 2.1 ViT-L & 0.288 & $-$0.068 & 3.001 & 0.498 & 0.089 & 2.830 \\
VideoMAEv2 giant & 0.292 & $-$0.044 & 2.864 & 0.437 & $-$0.080 & 3.068 \\
LeVJEPA ViT-L & 0.276 & $-$0.085 & 2.889 & 0.400 & $-$0.126 & 2.906 \\
VideoMAEv2 ViT-L & 0.174 & $-$0.100 & 2.866 & 0.354 & $-$0.053 & 2.832 \\
\midrule
\multicolumn{7}{@{}l}{\textit{Generative world models}} \\
DreamZero-DROID & 0.233 & $-$0.089 & 3.167 & 0.446 & $-$0.233 & 3.176 \\
Cosmos3-Nano & 0.045 & $-$0.293 & 3.168 & 0.363 & $-$0.051 & 2.849 \\
Cosmos3-Nano DROID & 0.013 & $-$0.263 & 3.115 & 0.316 & $-$0.137 & 3.152 \\
Cosmos-Predict2 2B & 0.102 & $-$0.163 & 2.810 & 0.327 & $-$0.206 & 2.969 \\
Cosmos3-Edge & $-$0.008 & $-$0.191 & 2.861 & 0.307 & $-$0.243 & 3.126 \\
Wan 2.1 I2V 14B & 0.188 & $-$0.129 & 2.943 & 0.312 & $-$0.122 & 3.042 \\
Cosmos3-Edge DROID & $-$0.072 & $-$0.256 & 3.163 & 0.255 & $-$0.143 & 2.911 \\
Wan 2.1 1.3B & 0.165 & $-$0.214 & 3.248 & 0.258 & $-$0.083 & 3.107 \\
\midrule
\multicolumn{7}{@{}l}{\textit{Vision--language / action models}} \\
Cosmos-Reason2 (16 blocks) & 0.054 & $-$0.116 & 3.064 & 0.303 & $-$0.037 & 2.830 \\
Cosmos-Reason2 (video) & 0.084 & $-$0.110 & 2.938 & 0.252 & $-$0.159 & 2.897 \\
Cosmos-Reason2 (image) & 0.045 & $-$0.101 & 3.024 & 0.255 & $-$0.062 & 2.856 \\
GR00T N1.7 (2 frames) & 0.052 & $-$0.101 & 2.949 & 0.257 & $-$0.146 & 2.773 \\
GR00T N1.7 (1 image) & 0.032 & $-$0.112 & 2.820 & 0.224 & $-$0.153 & 2.774 \\
$\pi_0$ & 0.140 & $-$0.211 & 3.022 & 0.139 & $-$0.157 & 3.007 \\
OpenVLA-7B & 0.095 & $-$0.031 & 2.979 & 0.101 & $-$0.064 & 2.759 \\
\bottomrule
\end{tabular}
}
\end{table}

%% file: tables/joint_selection/table_bootstrap_views.tex
\begin{table*}[tp]
\centering
\caption{Bootstrap uncertainty of the episode correlation on \textbf{Hoi!}. We report $r$ with its scene-cluster standard error, as in the main tables, and percentile 95\% intervals from resampling whole recordings (nine) or scenes (six) in 10,000 draws. Both choices give similar intervals.}
\label{tab:bootstrap}
\small
\setlength{\tabcolsep}{4pt}
\begin{tabular}{@{}lcccccc@{}}
\toprule
& \multicolumn{3}{c}{Human view} & \multicolumn{3}{c}{Gripper view} \\
\cmidrule(lr){2-4}\cmidrule(lr){5-7}
Representation & $r$ & Recording & Scene & $r$ & Recording & Scene \\
\midrule
\multicolumn{7}{@{}l}{\textit{Image and geometry encoders}} \\
DINOv3 ViT-H+ & $0.173\forcese{0.039}$ & $[0.094,\,0.240]$ & $[0.084,\,0.245]$ & $0.406\forcese{0.027}$ & $[0.353,\,0.463]$ & $[0.363,\,0.467]$ \\
VGGT-Omega 1B & $0.237\forcese{0.022}$ & $[0.199,\,0.275]$ & $[0.194,\,0.281]$ & $0.400\forcese{0.034}$ & $[0.339,\,0.466]$ & $[0.343,\,0.477]$ \\
DINOv3 +diff & $0.155\forcese{0.027}$ & $[0.097,\,0.210]$ & $[0.095,\,0.198]$ & $0.321\forcese{0.048}$ & $[0.239,\,0.393]$ & $[0.215,\,0.401]$ \\
\midrule
\multicolumn{7}{@{}l}{\textit{Video encoders}} \\
V-JEPA 2.1 ViT-G & $0.337\forcese{0.048}$ & $[0.253,\,0.432]$ & $[0.254,\,0.446]$ & $0.527\forcese{0.019}$ & $[0.488,\,0.559]$ & $[0.483,\,0.557]$ \\
V-JEPA 2 ViT-L & $0.313\forcese{0.038}$ & $[0.234,\,0.379]$ & $[0.238,\,0.384]$ & $0.511\forcese{0.029}$ & $[0.463,\,0.559]$ & $[0.454,\,0.571]$ \\
V-JEPA 2.1 ViT-L & $0.342\forcese{0.034}$ & $[0.282,\,0.405]$ & $[0.284,\,0.416]$ & $0.496\forcese{0.014}$ & $[0.469,\,0.518]$ & $[0.464,\,0.520]$ \\
VideoMAEv2 giant & $0.348\forcese{0.027}$ & $[0.297,\,0.400]$ & $[0.296,\,0.402]$ & $0.480\forcese{0.040}$ & $[0.408,\,0.540]$ & $[0.393,\,0.549]$ \\
LeVJEPA ViT-L & $0.278\forcese{0.033}$ & $[0.217,\,0.329]$ & $[0.200,\,0.331]$ & $0.444\forcese{0.027}$ & $[0.383,\,0.501]$ & $[0.390,\,0.492]$ \\
VideoMAEv2 ViT-L & $0.173\forcese{0.023}$ & $[0.128,\,0.209]$ & $[0.115,\,0.208]$ & $0.414\forcese{0.032}$ & $[0.355,\,0.473]$ & $[0.353,\,0.478]$ \\
\midrule
\multicolumn{7}{@{}l}{\textit{Generative world models}} \\
DreamZero-DROID & $0.279\forcese{0.040}$ & $[0.212,\,0.367]$ & $[0.225,\,0.376]$ & $0.458\forcese{0.037}$ & $[0.380,\,0.526]$ & $[0.378,\,0.521]$ \\
Cosmos3-Nano & $0.114\forcese{0.057}$ & $[0.016,\,0.204]$ & $[-0.003,\,0.220]$ & $0.419\forcese{0.029}$ & $[0.364,\,0.477]$ & $[0.371,\,0.484]$ \\
Cosmos3-Nano DROID & $0.084\forcese{0.030}$ & $[0.026,\,0.150]$ & $[0.025,\,0.143]$ & $0.383\forcese{0.053}$ & $[0.293,\,0.458]$ & $[0.277,\,0.488]$ \\
Cosmos-Predict2 2B & $0.159\forcese{0.036}$ & $[0.085,\,0.225]$ & $[0.086,\,0.226]$ & $0.368\forcese{0.045}$ & $[0.278,\,0.442]$ & $[0.272,\,0.447]$ \\
Cosmos3-Edge & $0.110\forcese{0.056}$ & $[-0.001,\,0.216]$ & $[0.003,\,0.219]$ & $0.358\forcese{0.051}$ & $[0.267,\,0.416]$ & $[0.242,\,0.421]$ \\
Wan 2.1 I2V 14B & $0.225\forcese{0.036}$ & $[0.165,\,0.293]$ & $[0.162,\,0.308]$ & $0.353\forcese{0.023}$ & $[0.310,\,0.396]$ & $[0.300,\,0.389]$ \\
Cosmos3-Edge DROID & $0.071\forcese{0.066}$ & $[-0.051,\,0.188]$ & $[-0.057,\,0.207]$ & $0.338\forcese{0.047}$ & $[0.253,\,0.400]$ & $[0.231,\,0.418]$ \\
Wan 2.1 1.3B & $0.250\forcese{0.049}$ & $[0.166,\,0.329]$ & $[0.146,\,0.339]$ & $0.333\forcese{0.034}$ & $[0.273,\,0.395]$ & $[0.267,\,0.397]$ \\
\midrule
\multicolumn{7}{@{}l}{\textit{Vision--language / action models}} \\
Cosmos-Reason2 (16 blocks) & $0.102\forcese{0.024}$ & $[0.043,\,0.155]$ & $[0.054,\,0.145]$ & $0.335\forcese{0.030}$ & $[0.266,\,0.408]$ & $[0.270,\,0.387]$ \\
Cosmos-Reason2 (video) & $0.155\forcese{0.049}$ & $[0.073,\,0.243]$ & $[0.066,\,0.266]$ & $0.327\forcese{0.046}$ & $[0.245,\,0.415]$ & $[0.249,\,0.428]$ \\
Cosmos-Reason2 (image) & $0.102\forcese{0.033}$ & $[0.026,\,0.182]$ & $[0.041,\,0.172]$ & $0.310\forcese{0.037}$ & $[0.241,\,0.385]$ & $[0.232,\,0.375]$ \\
GR00T N1.7 (2 frames) & $0.137\forcese{0.041}$ & $[0.049,\,0.221]$ & $[0.056,\,0.217]$ & $0.300\forcese{0.027}$ & $[0.251,\,0.357]$ & $[0.254,\,0.357]$ \\
GR00T N1.7 (1 image) & $0.101\forcese{0.039}$ & $[0.025,\,0.192]$ & $[0.040,\,0.191]$ & $0.267\forcese{0.028}$ & $[0.217,\,0.329]$ & $[0.226,\,0.335]$ \\
$\pi_0$ & $0.150\forcese{0.030}$ & $[0.090,\,0.193]$ & $[0.074,\,0.195]$ & $0.231\forcese{0.044}$ & $[0.140,\,0.319]$ & $[0.139,\,0.315]$ \\
OpenVLA-7B & $0.094\forcese{0.010}$ & $[0.066,\,0.120]$ & $[0.075,\,0.116]$ & $0.152\forcese{0.024}$ & $[0.075,\,0.229]$ & $[0.108,\,0.207]$ \\
\bottomrule
\end{tabular}
\end{table*}

%% file: tables/joint_selection/table_peaks_views.tex
\begin{table*}[tb]
\centering
\caption{Peak timing and magnitude on \textbf{Hoi!} (131 human-view and 132 gripper-view contact episodes). We compare the predicted and measured maximum of each episode and report the median timing offset $|\Delta t|$ (s), the share of peaks (\%) within one or two time steps (0.25\,s each), and the error of the peak force (N). Video encoders locate the peak most often, but no model does so reliably.}
\label{tab:peak_diagnostics}
\footnotesize
\setlength{\tabcolsep}{3pt}
\begin{tabular}{@{}lrrrrrrrrrrrr@{}}
\toprule
& \multicolumn{6}{c}{Human view} & \multicolumn{6}{c}{Gripper view} \\
\cmidrule(lr){2-7}\cmidrule(lr){8-13}
& \multicolumn{3}{c}{Timing} & \multicolumn{3}{c}{Magnitude} & \multicolumn{3}{c}{Timing} & \multicolumn{3}{c}{Magnitude} \\
\cmidrule(lr){2-4}\cmidrule(lr){5-7}\cmidrule(lr){8-10}\cmidrule(lr){11-13}
Representation & $|\Delta t|$ & $\leq$1 & $\leq$2 & MAE & MedAE & RMSE & $|\Delta t|$ & $\leq$1 & $\leq$2 & MAE & MedAE & RMSE \\
\midrule
\multicolumn{13}{@{}l}{\textit{Image and geometry encoders}} \\
DINOv3 ViT-H+ & 0.50 & 31.3 & 51.9 & 5.11 & 2.51 & 8.31 & 0.50 & 35.6 & 50.8 & 4.50 & 2.47 & 7.47 \\
VGGT-Omega 1B & 0.50 & 35.1 & 52.7 & 4.76 & 2.77 & 7.38 & 0.63 & 36.4 & 50.0 & 5.00 & 2.83 & 8.15 \\
DINOv3 +diff & 0.75 & 29.0 & 44.3 & 4.94 & 2.80 & 7.81 & 0.75 & 34.8 & 44.7 & 5.53 & 3.44 & 8.78 \\
\midrule
\multicolumn{13}{@{}l}{\textit{Video encoders}} \\
V-JEPA 2.1 ViT-G & 0.25 & 53.4 & 64.1 & 5.74 & 4.07 & 7.67 & 0.50 & 45.5 & 60.6 & 4.49 & 2.93 & 7.10 \\
V-JEPA 2 ViT-L & 0.50 & 40.5 & 54.2 & 6.51 & 4.57 & 8.98 & 0.50 & 48.5 & 61.4 & 4.26 & 2.87 & 6.85 \\
V-JEPA 2.1 ViT-L & 0.50 & 48.9 & 65.6 & 5.19 & 3.52 & 7.67 & 0.25 & 50.8 & 60.6 & 4.02 & 3.09 & 6.30 \\
VideoMAEv2 giant & 0.50 & 47.3 & 57.3 & 4.79 & 2.82 & 7.67 & 0.50 & 37.9 & 54.5 & 4.66 & 3.11 & 7.33 \\
LeVJEPA ViT-L & 0.50 & 42.7 & 53.4 & 4.58 & 2.71 & 7.18 & 0.50 & 46.2 & 63.6 & 4.77 & 2.70 & 7.79 \\
VideoMAEv2 ViT-L & 0.75 & 28.2 & 42.7 & 4.96 & 2.81 & 8.04 & 0.50 & 36.4 & 56.1 & 4.66 & 2.66 & 7.70 \\
\midrule
\multicolumn{13}{@{}l}{\textit{Generative world models}} \\
DreamZero-DROID & 0.50 & 39.7 & 58.8 & 4.37 & 3.03 & 6.96 & 0.50 & 42.4 & 56.8 & 4.83 & 3.60 & 7.22 \\
Cosmos3-Nano & 0.50 & 31.3 & 50.4 & 5.64 & 3.36 & 8.85 & 0.50 & 43.9 & 58.3 & 4.95 & 3.17 & 7.97 \\
Cosmos3-Nano DROID & 0.50 & 31.3 & 51.1 & 5.69 & 3.39 & 9.02 & 0.63 & 35.6 & 50.0 & 4.70 & 2.52 & 7.95 \\
Cosmos-Predict2 2B & 0.75 & 34.4 & 48.1 & 5.84 & 3.54 & 9.06 & 0.50 & 40.9 & 62.1 & 5.63 & 3.80 & 8.11 \\
Cosmos3-Edge & 1.00 & 22.9 & 39.7 & 5.86 & 3.09 & 9.20 & 0.75 & 34.8 & 47.0 & 5.04 & 2.69 & 8.28 \\
Wan 2.1 I2V 14B & 0.75 & 38.9 & 49.6 & 5.24 & 3.08 & 8.43 & 0.50 & 37.1 & 53.8 & 5.46 & 3.17 & 8.73 \\
Cosmos3-Edge DROID & 0.75 & 29.8 & 44.3 & 5.47 & 2.80 & 8.79 & 0.75 & 31.8 & 46.2 & 5.25 & 2.89 & 8.55 \\
Wan 2.1 1.3B & 0.75 & 32.8 & 44.3 & 5.37 & 3.35 & 8.57 & 0.50 & 34.8 & 56.8 & 4.60 & 2.97 & 7.37 \\
\midrule
\multicolumn{13}{@{}l}{\textit{Vision--language / action models}} \\
Cosmos-Reason2 (16 blocks) & 0.75 & 28.2 & 42.0 & 4.65 & 2.77 & 7.23 & 0.50 & 35.6 & 52.3 & 4.84 & 2.48 & 8.09 \\
Cosmos-Reason2 (video) & 0.75 & 31.3 & 47.3 & 4.90 & 2.67 & 7.79 & 0.50 & 37.9 & 52.3 & 5.13 & 3.24 & 7.88 \\
Cosmos-Reason2 (image) & 0.75 & 32.1 & 46.6 & 4.54 & 2.63 & 7.17 & 0.75 & 38.6 & 48.5 & 4.86 & 2.49 & 8.02 \\
GR00T N1.7 (2 frames) & 0.50 & 35.9 & 54.2 & 4.68 & 3.07 & 7.40 & 0.50 & 31.1 & 53.0 & 4.81 & 2.92 & 7.89 \\
GR00T N1.7 (1 image) & 0.75 & 25.2 & 42.7 & 4.81 & 2.79 & 7.79 & 0.63 & 34.8 & 50.0 & 4.71 & 2.39 & 7.81 \\
$\pi_0$ & 0.75 & 22.1 & 44.3 & 5.06 & 2.43 & 8.41 & 0.50 & 30.3 & 50.8 & 5.13 & 3.44 & 8.07 \\
OpenVLA-7B & 0.75 & 35.1 & 48.1 & 5.36 & 4.12 & 7.24 & 0.75 & 25.0 & 43.9 & 4.85 & 2.44 & 7.96 \\
\midrule
Uniform peak time & -- & 22.4 & 37.3 & -- & -- & -- & -- & 22.7 & 37.9 & -- & -- & -- \\
Median training peak & -- & -- & -- & 6.68 & 6.71 & 7.91 & -- & -- & -- & 6.67 & 6.68 & 7.91 \\
\bottomrule
\end{tabular}
\end{table*}

%% file: tables/joint_selection/table_range_bins_views.tex
\begin{table*}[tp]
\centering
\caption{Prediction range and force-binned error on \textbf{Hoi!}. Ratio: mean within-episode SD of the prediction over that of the measured force. Max and $>$12\,N: largest prediction (N) and number of time steps predicted above 12\,N. MAE (N) is computed over all evaluation time steps, binned by percentiles of the training force.}
\label{tab:range_bins}
\footnotesize
\setlength{\tabcolsep}{2.5pt}
\begin{tabular}{@{}lrrrrrrrrrrrrrr@{}}
\toprule
& \multicolumn{7}{c}{Human view} & \multicolumn{7}{c}{Gripper view} \\
\cmidrule(lr){2-8}\cmidrule(lr){9-15}
& \multicolumn{3}{c}{Range} & \multicolumn{4}{c}{MAE by percentile} & \multicolumn{3}{c}{Range} & \multicolumn{4}{c}{MAE by percentile} \\
\cmidrule(lr){2-4}\cmidrule(lr){5-8}\cmidrule(lr){9-11}\cmidrule(lr){12-15}
Representation & Ratio & Max & $>$12 & 0--50 & 50--75 & 75--90 & 90+ & Ratio & Max & $>$12 & 0--50 & 50--75 & 75--90 & 90+ \\
\midrule
\multicolumn{15}{@{}l}{\textit{Image and geometry encoders}} \\
DINOv3 ViT-H+ & 0.30 & 7.2 & 0 & 1.04 & 1.11 & 1.74 & 8.41 & 0.43 & 9.5 & 0 & 1.12 & 1.79 & 2.07 & 6.83 \\
VGGT-Omega 1B & 0.54 & 12.9 & 3 & 0.93 & 1.62 & 2.36 & 7.57 & 0.38 & 9.0 & 0 & 0.81 & 1.53 & 1.73 & 7.78 \\
DINOv3 +diff & 0.39 & 8.7 & 0 & 1.09 & 1.15 & 1.90 & 8.60 & 0.26 & 6.9 & 0 & 1.08 & 0.89 & 1.90 & 8.84 \\
\midrule
\multicolumn{15}{@{}l}{\textit{Video encoders}} \\
V-JEPA 2.1 ViT-G & 1.02 & 27.8 & 74 & 1.65 & 1.84 & 2.60 & 6.25 & 0.72 & 12.7 & 2 & 0.88 & 1.45 & 1.77 & 6.84 \\
V-JEPA 2 ViT-L & 1.10 & 26.6 & 121 & 1.07 & 1.35 & 3.14 & 7.48 & 0.83 & 15.5 & 7 & 0.83 & 1.48 & 2.27 & 6.36 \\
V-JEPA 2.1 ViT-L & 0.75 & 18.7 & 47 & 1.47 & 1.85 & 2.31 & 6.97 & 0.80 & 15.2 & 14 & 0.89 & 1.45 & 2.13 & 5.89 \\
VideoMAEv2 giant & 0.44 & 11.7 & 0 & 1.38 & 1.52 & 1.93 & 7.59 & 0.66 & 10.2 & 0 & 1.08 & 1.72 & 2.32 & 6.82 \\
LeVJEPA ViT-L & 0.55 & 13.7 & 8 & 1.33 & 1.49 & 2.01 & 7.44 & 0.60 & 12.7 & 2 & 0.67 & 1.00 & 2.00 & 8.52 \\
VideoMAEv2 ViT-L & 0.36 & 10.9 & 0 & 1.62 & 1.51 & 1.95 & 7.84 & 0.54 & 9.7 & 0 & 1.17 & 1.51 & 1.75 & 7.71 \\
\midrule
\multicolumn{15}{@{}l}{\textit{Generative world models}} \\
DreamZero-DROID & 0.66 & 13.1 & 5 & 1.26 & 1.67 & 2.28 & 6.75 & 0.82 & 21.4 & 37 & 1.01 & 1.58 & 2.64 & 6.82 \\
Cosmos3-Nano & 0.34 & 10.3 & 0 & 1.53 & 1.27 & 2.26 & 9.36 & 0.49 & 10.2 & 0 & 1.73 & 1.56 & 1.83 & 7.46 \\
Cosmos3-Nano DROID & 0.30 & 8.8 & 0 & 1.43 & 1.15 & 2.26 & 9.62 & 0.47 & 9.2 & 0 & 1.47 & 1.46 & 1.87 & 7.90 \\
Cosmos-Predict2 2B & 0.45 & 13.2 & 3 & 0.82 & 1.21 & 2.75 & 9.91 & 0.87 & 19.1 & 112 & 1.34 & 2.12 & 3.18 & 7.68 \\
Cosmos3-Edge & 0.24 & 6.3 & 0 & 1.45 & 1.08 & 2.06 & 9.70 & 0.41 & 9.0 & 0 & 1.32 & 1.29 & 1.77 & 8.37 \\
Wan 2.1 I2V 14B & 0.32 & 8.0 & 0 & 1.08 & 1.27 & 2.00 & 8.54 & 0.48 & 11.3 & 0 & 0.83 & 1.22 & 1.99 & 9.30 \\
Cosmos3-Edge DROID & 0.24 & 6.0 & 0 & 1.65 & 1.06 & 1.80 & 9.20 & 0.28 & 6.5 & 0 & 1.40 & 1.08 & 1.60 & 8.59 \\
Wan 2.1 1.3B & 0.47 & 10.1 & 0 & 1.46 & 1.35 & 2.35 & 8.93 & 0.70 & 12.1 & 1 & 0.95 & 1.42 & 2.59 & 8.11 \\
\midrule
\multicolumn{15}{@{}l}{\textit{Vision--language / action models}} \\
Cosmos-Reason2 (16 blocks) & 0.40 & 10.8 & 0 & 1.48 & 1.84 & 1.74 & 7.61 & 0.47 & 10.2 & 0 & 0.62 & 1.28 & 1.77 & 8.42 \\
Cosmos-Reason2 (video) & 0.45 & 10.8 & 0 & 0.98 & 1.35 & 2.00 & 8.35 & 0.73 & 15.0 & 28 & 0.65 & 1.63 & 2.31 & 8.04 \\
Cosmos-Reason2 (image) & 0.49 & 13.0 & 4 & 1.10 & 1.48 & 1.93 & 8.13 & 0.41 & 9.5 & 0 & 0.78 & 1.28 & 1.72 & 8.38 \\
GR00T N1.7 (2 frames) & 0.47 & 11.8 & 0 & 0.83 & 1.55 & 2.08 & 7.95 & 0.47 & 9.8 & 0 & 0.83 & 1.39 & 2.07 & 8.04 \\
GR00T N1.7 (1 image) & 0.35 & 10.0 & 0 & 1.12 & 1.35 & 1.74 & 8.15 & 0.45 & 9.3 & 0 & 1.09 & 1.27 & 1.74 & 8.39 \\
$\pi_0$ & 0.15 & 6.2 & 0 & 1.56 & 1.65 & 1.47 & 8.12 & 0.52 & 10.4 & 0 & 0.74 & 1.33 & 1.92 & 8.71 \\
OpenVLA-7B & 0.83 & 18.7 & 35 & 2.24 & 2.51 & 2.41 & 7.39 & 0.39 & 9.8 & 0 & 1.33 & 1.21 & 1.58 & 8.64 \\
\midrule
Median training force & 0.00 & -- & -- & 0.61 & 0.46 & 3.02 & 10.92 & 0.00 & -- & -- & 0.61 & 0.46 & 3.02 & 10.91 \\
\bottomrule
\end{tabular}
\end{table*}

%% file: tables/rh20t/table_rh20t_r_by_peak.tex
\begin{table}[tp]
\centering
\caption{Episode correlation on \textbf{RH20T}. We report $r$ with its 95\% interval from the operator--task bootstrap and $r$ by the measured peak force of the episode. Below 5\,N, which covers 45\% of the episodes, no model tracks force.}
\label{tab:rh20t_r_by_peak}
\footnotesize
\setlength{\tabcolsep}{2.5pt}
\resizebox{\columnwidth}{!}{%
\begin{tabular}{@{}lrcrrrr@{}}
\toprule
& \multicolumn{2}{c}{All episodes} & \multicolumn{4}{c}{By episode peak (N)} \\
\cmidrule(lr){2-3}\cmidrule(lr){4-7}
Representation & $r$ & 95\% interval & 2--5 & 5--10 & 10--20 & $\geq$20 \\
\textit{Episodes} & & & \textit{2998} & \textit{1245} & \textit{1086} & \textit{1271} \\
\midrule
\multicolumn{7}{@{}l}{\textit{Video-free reference}} \\
Task prior (best of 3) & 0.023 & [0.014, 0.033] & 0.01 & 0.03 & 0.05 & 0.03 \\
\midrule
\multicolumn{7}{@{}l}{\textit{Image and geometry encoders}} \\
DINOv3 ViT-H+ & 0.081 & [0.066, 0.098] & 0.03 & 0.09 & 0.15 & 0.14 \\
VGGT-Omega 1B & 0.077 & [0.061, 0.093] & 0.02 & 0.08 & 0.13 & 0.16 \\
\midrule
\multicolumn{7}{@{}l}{\textit{Video encoders}} \\
V-JEPA 2.1 ViT-G & 0.203 & [0.167, 0.242] & 0.03 & 0.20 & 0.37 & 0.48 \\
VideoMAEv2 giant & 0.192 & [0.156, 0.230] & 0.04 & 0.18 & 0.35 & 0.43 \\
\midrule
\multicolumn{7}{@{}l}{\textit{Generative world models}} \\
Wan 2.1 I2V 14B & 0.184 & [0.147, 0.222] & 0.02 & 0.16 & 0.32 & 0.48 \\
DreamZero-DROID & 0.150 & [0.120, 0.182] & 0.01 & 0.13 & 0.26 & 0.40 \\
Cosmos3-Edge & 0.134 & [0.110, 0.159] & 0.02 & 0.14 & 0.25 & 0.30 \\
Cosmos3-Edge DROID & 0.132 & [0.107, 0.158] & 0.01 & 0.12 & 0.25 & 0.33 \\
\midrule
\multicolumn{7}{@{}l}{\textit{Vision--language / action models}} \\
$\pi_0$ & 0.069 & [0.051, 0.088] & 0.01 & 0.09 & 0.14 & 0.14 \\
GR00T N1.7 (2 frames) & 0.032 & [0.022, 0.043] & 0.01 & 0.02 & 0.07 & 0.07 \\
\bottomrule
\end{tabular}%
}
\end{table}

%% file: tables/rh20t/table_rh20t_peaks.tex
\begin{table*}[p]
\centering
\caption{Peak timing, peak magnitude, and prediction range on \textbf{RH20T} (6600 contact episodes), as in Tables~\ref{tab:peak_diagnostics} and~\ref{tab:range_bins}. Ratio: mean within-episode SD of the prediction over that of the measured force. Max: largest prediction (N). MAE (N) by percentile is computed over all evaluation time steps, binned by percentiles of the training force (1.2, 2.3, and 6.1\,N).}
\label{tab:rh20t_peaks}
\footnotesize
\setlength{\tabcolsep}{3pt}
\begin{tabular}{@{}lrrrrrrrrrrrr@{}}
\toprule
& \multicolumn{3}{c}{Peak timing} & \multicolumn{3}{c}{Peak magnitude} & \multicolumn{2}{c}{Range} & \multicolumn{4}{c}{MAE by percentile} \\
\cmidrule(lr){2-4}\cmidrule(lr){5-7}\cmidrule(lr){8-9}\cmidrule(lr){10-13}
Representation & $|\Delta t|$ & $\leq$1 & $\leq$2 & MAE & MedAE & RMSE & Ratio & Max & 0--50 & 50--75 & 75--90 & 90+ \\
\midrule
\multicolumn{13}{@{}l}{\textit{Video-free reference}} \\
Task prior (best of 3) & 1.00 & 16.7 & 34.0 & 11.73 & 3.82 & 27.51 & 0.29 & 133.4 & 2.26 & 3.06 & 4.06 & 14.09 \\
\midrule
\multicolumn{13}{@{}l}{\textit{Image and geometry encoders}} \\
DINOv3 ViT-H+ & 0.75 & 26.3 & 44.4 & 10.70 & 3.66 & 26.33 & 0.36 & 27.6 & 2.38 & 2.38 & 2.80 & 10.87 \\
VGGT-Omega 1B & 0.75 & 26.2 & 44.4 & 11.22 & 3.53 & 27.03 & 0.31 & 26.5 & 1.66 & 1.87 & 2.17 & 12.41 \\
\midrule
\multicolumn{13}{@{}l}{\textit{Video encoders}} \\
V-JEPA 2.1 ViT-G & 0.50 & 36.4 & 51.5 & 9.75 & 4.41 & 24.70 & 0.71 & 42.5 & 2.06 & 2.44 & 3.05 & 9.62 \\
VideoMAEv2 giant & 0.50 & 34.7 & 50.7 & 10.35 & 3.15 & 26.36 & 0.37 & 24.0 & 1.64 & 1.66 & 1.90 & 11.30 \\
\midrule
\multicolumn{13}{@{}l}{\textit{Generative world models}} \\
Wan 2.1 I2V 14B & 0.50 & 34.3 & 50.2 & 9.76 & 3.70 & 25.35 & 0.53 & 40.0 & 1.73 & 1.92 & 2.29 & 10.71 \\
DreamZero-DROID & 0.75 & 33.3 & 47.9 & 9.73 & 4.24 & 24.45 & 0.69 & 55.5 & 2.01 & 2.28 & 3.06 & 10.33 \\
Cosmos3-Edge & 0.75 & 30.5 & 47.1 & 10.46 & 3.25 & 26.45 & 0.36 & 23.4 & 2.03 & 1.92 & 2.14 & 11.42 \\
Cosmos3-Edge DROID & 0.75 & 33.1 & 48.9 & 10.52 & 3.10 & 26.56 & 0.34 & 20.2 & 1.55 & 1.63 & 1.94 & 11.98 \\
\midrule
\multicolumn{13}{@{}l}{\textit{Vision--language / action models}} \\
$\pi_0$ & 0.75 & 24.3 & 40.5 & 11.26 & 2.95 & 27.57 & 0.17 & 39.1 & 2.13 & 1.98 & 1.99 & 11.72 \\
GR00T N1.7 (2 frames) & 0.75 & 25.2 & 42.5 & 11.07 & 2.92 & 27.28 & 0.18 & 29.2 & 2.10 & 1.45 & 1.46 & 12.70 \\
\midrule
Uniform peak time & -- & 25.3 & 42.1 & -- & -- & -- & -- & -- & -- & -- & -- & -- \\
Median training peak & -- & -- & -- & 11.34 & 2.84 & 27.94 & -- & -- & -- & -- & -- & -- \\
Median training force & -- & -- & -- & -- & -- & -- & 0 & -- & 0.53 & 0.47 & 2.34 & 16.53 \\
\bottomrule
\end{tabular}
\end{table*}

%% file: figures/vlm/figure_vlm_peak_scatter.tex
\begin{figure*}[t]
\centering
\includegraphics[width=\textwidth]{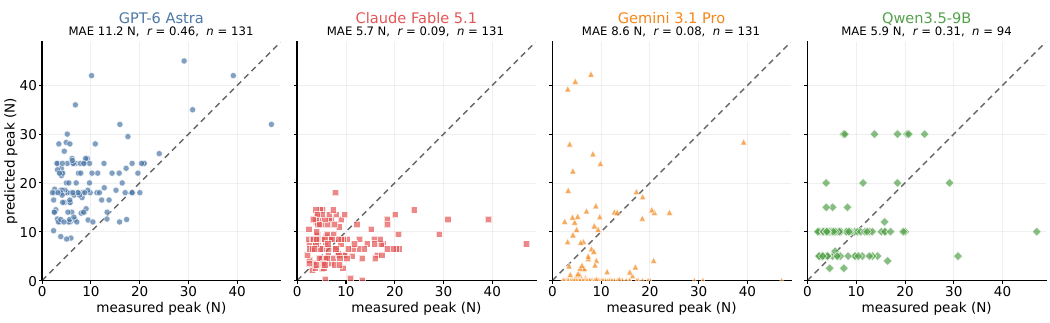}
\caption{Zero-shot peak force from multimodal LLMs on the 131 \textbf{Hoi!} human-view contact episodes. Each model sees 16 frames of an episode and is asked for its peak contact force in newtons. Ground truth (x-axis) is the measured peak of the episode, the maximum of the per-step force target used by the probes. The dashed line marks a correct estimate, and $n$ counts parseable answers. No model estimates force reliably; predicting the median training peak gives an MAE of 6.7\,N.}
\label{fig:vlm_peak}
\end{figure*}

%% file: tables/strict_contact/table_strict_contact.tex
\begin{table*}[p]
\centering
\caption{Strict-contact evaluation on \textbf{Hoi!} and \textbf{RH20T}. \textit{Padded}: contact episodes as in the main paper. \textit{Strict}: only the time steps at or above 2\,N within them. Probes and locations are those of the main tables. Correlation barely changes on RH20T and in the Hoi!\ human view, and roughly halves in the Hoi!\ gripper view.}
\label{tab:strict_contact}
\footnotesize
\setlength{\tabcolsep}{3pt}
\begin{tabular}{@{}lcccccccccccc@{}}
\toprule
& \multicolumn{4}{c}{Hoi! human view} & \multicolumn{4}{c}{Hoi! gripper view} & \multicolumn{4}{c}{RH20T} \\
\cmidrule(lr){2-5}\cmidrule(lr){6-9}\cmidrule(lr){10-13}
& \multicolumn{2}{c}{$r\uparrow$} & \multicolumn{2}{c}{$R^2\uparrow$} & \multicolumn{2}{c}{$r\uparrow$} & \multicolumn{2}{c}{$R^2\uparrow$} & \multicolumn{2}{c}{$r\uparrow$} & \multicolumn{2}{c}{$R^2\uparrow$} \\
Representation & padded & strict & padded & strict & padded & strict & padded & strict & padded & strict & padded & strict \\
\midrule
\multicolumn{13}{@{}l}{\textit{Image and geometry encoders}} \\
DINOv3 ViT-H+ & $0.173$ & $0.166$ & $-0.033$ & $-0.217$ & $0.406$ & $0.132$ & $0.008$ & $-0.136$ & $0.081$ & $0.079$ & $0.037$ & $-0.007$ \\
VGGT-Omega 1B & $0.237$ & $0.222$ & $-0.047$ & $-0.213$ & $0.400$ & $0.184$ & $0.018$ & $-0.121$ & $0.077$ & $0.076$ & $0.058$ & $0.019$ \\
\midrule
\multicolumn{13}{@{}l}{\textit{Video encoders}} \\
V-JEPA 2.1 ViT-G & $0.337$ & $0.335$ & $-0.055$ & $-0.177$ & $0.527$ & $0.272$ & $0.166$ & $0.052$ & $0.203$ & $0.195$ & $0.133$ & $0.105$ \\
VideoMAEv2 giant & $0.348$ & $0.273$ & $-0.037$ & $-0.181$ & $0.480$ & $0.262$ & $-0.012$ & $-0.163$ & $0.192$ & $0.167$ & $0.071$ & $0.022$ \\
\midrule
\multicolumn{13}{@{}l}{\textit{Generative world models}} \\
DreamZero-DROID & $0.279$ & $0.197$ & $-0.069$ & $-0.118$ & $0.458$ & $0.331$ & $-0.118$ & $-0.308$ & $0.150$ & $0.147$ & $0.081$ & $0.033$ \\
Cosmos3-Nano & $0.114$ & $0.107$ & $-0.297$ & $-0.579$ & $0.419$ & $0.220$ & $-0.037$ & $-0.176$ & -- & -- & -- & -- \\
Cosmos3-Nano DROID & $0.084$ & $0.116$ & $-0.263$ & $-0.541$ & $0.383$ & $0.163$ & $-0.063$ & $-0.181$ & -- & -- & -- & -- \\
Cosmos-Predict2 2B & $0.159$ & $0.164$ & $-0.168$ & $-0.460$ & $0.368$ & $0.211$ & $-0.146$ & $-0.409$ & -- & -- & -- & -- \\
Cosmos3-Edge & $0.110$ & $0.143$ & $-0.192$ & $-0.472$ & $0.358$ & $0.203$ & $-0.204$ & $-0.467$ & $0.134$ & $0.124$ & $0.043$ & $0.000$ \\
Wan 2.1 I2V 14B & $0.225$ & $0.153$ & $-0.097$ & $-0.315$ & $0.353$ & $0.202$ & $-0.064$ & $-0.263$ & $0.184$ & $0.179$ & $0.073$ & $0.024$ \\
Cosmos3-Edge DROID & $0.071$ & $0.097$ & $-0.215$ & $-0.437$ & $0.338$ & $0.136$ & $-0.086$ & $-0.293$ & $0.132$ & $0.127$ & $0.032$ & $-0.019$ \\
Wan 2.1 1.3B & $0.250$ & $0.207$ & $-0.177$ & $-0.318$ & $0.333$ & $0.236$ & $-0.047$ & $-0.204$ & -- & -- & -- & -- \\
\midrule
\multicolumn{13}{@{}l}{\textit{Vision--language / action models}} \\
Cosmos-Reason2 (video) & $0.155$ & $0.044$ & $-0.125$ & $-0.301$ & $0.327$ & $0.141$ & $-0.117$ & $-0.327$ & -- & -- & -- & -- \\
GR00T N1.7 (2 frames) & $0.137$ & $0.103$ & $-0.087$ & $-0.264$ & $0.300$ & $0.162$ & $-0.129$ & $-0.377$ & $0.032$ & $0.041$ & $0.020$ & $-0.024$ \\
$\pi_0$ & $0.150$ & $0.147$ & $-0.213$ & $-0.405$ & $0.231$ & $0.098$ & $-0.098$ & $-0.298$ & $0.069$ & $0.074$ & $-0.014$ & $-0.071$ \\
OpenVLA-7B & $0.094$ & $0.166$ & $-0.080$ & $-0.194$ & $0.152$ & $0.061$ & $-0.061$ & $-0.243$ & -- & -- & -- & -- \\
\midrule
V-JEPA 2.1 ViT-G + LoRA & -- & -- & -- & -- & -- & -- & -- & -- & $0.325$ & $0.316$ & $0.234$ & $0.195$ \\
\bottomrule
\end{tabular}
\end{table*}

%% file: tables/strict_contact/table_calibration_bound.tex
\begin{table*}[p]
\centering
\caption{Calibration bound on contact $R^2$. \textit{Cal.}: contact $R^2$ after the best affine rescaling $a\,\hat{y}+b$ of the probe output, fitted on the evaluated contact time steps. $a$: fitted slope. Even after calibration, contact $R^2$ stays low for every model.}
\label{tab:calibration_bound}
\footnotesize
\setlength{\tabcolsep}{4pt}
\begin{tabular}{@{}lccccccccc@{}}
\toprule
& \multicolumn{3}{c}{Hoi! human view} & \multicolumn{3}{c}{Hoi! gripper view} & \multicolumn{3}{c}{RH20T} \\
\cmidrule(lr){2-4}\cmidrule(lr){5-7}\cmidrule(lr){8-10}
Representation & $R^2\uparrow$ & Cal.$\uparrow$ & $a$ & $R^2\uparrow$ & Cal.$\uparrow$ & $a$ & $R^2\uparrow$ & Cal.$\uparrow$ & $a$ \\
\midrule
\multicolumn{10}{@{}l}{\textit{Image and geometry encoders}} \\
DINOv3 ViT-H+ & $-0.033$ & $0.024$ & $0.50$ & $0.008$ & $0.031$ & $0.56$ & $0.037$ & $0.059$ & $0.88$ \\
VGGT-Omega 1B & $-0.047$ & $0.011$ & $0.36$ & $0.018$ & $0.036$ & $0.60$ & $0.058$ & $0.068$ & $0.96$ \\
\midrule
\multicolumn{10}{@{}l}{\textit{Video encoders}} \\
V-JEPA 2.1 ViT-G & $-0.055$ & $0.034$ & $0.39$ & $0.166$ & $0.190$ & $0.74$ & $0.133$ & $0.137$ & $0.90$ \\
VideoMAEv2 giant & $-0.037$ & $0.025$ & $0.41$ & $-0.012$ & $0.048$ & $0.48$ & $0.071$ & $0.120$ & $1.28$ \\
\midrule
\multicolumn{10}{@{}l}{\textit{Generative world models}} \\
DreamZero-DROID & $-0.069$ & $0.058$ & $0.40$ & $-0.118$ & $0.059$ & $0.39$ & $0.081$ & $0.134$ & $1.03$ \\
Cosmos3-Nano & $-0.297$ & $0.002$ & $-0.13$ & $-0.037$ & $0.042$ & $0.42$ & -- & -- & -- \\
Cosmos3-Nano DROID & $-0.263$ & $0.001$ & $-0.08$ & $-0.063$ & $0.048$ & $0.40$ & -- & -- & -- \\
Cosmos-Predict2 2B & $-0.168$ & $0.010$ & $0.36$ & $-0.146$ & $0.028$ & $0.39$ & -- & -- & -- \\
Cosmos3-Edge & $-0.192$ & $0.003$ & $-0.20$ & $-0.204$ & $0.006$ & $0.17$ & $0.043$ & $0.061$ & $1.10$ \\
Wan 2.1 I2V 14B & $-0.097$ & $0.018$ & $0.37$ & $-0.064$ & $0.053$ & $0.45$ & $0.073$ & $0.126$ & $1.13$ \\
Cosmos3-Edge DROID & $-0.215$ & $0.002$ & $0.10$ & $-0.086$ & $0.006$ & $0.24$ & $0.032$ & $0.081$ & $1.29$ \\
Wan 2.1 1.3B & $-0.177$ & $0.028$ & $0.28$ & $-0.047$ & $0.032$ & $0.41$ & -- & -- & -- \\
\midrule
\multicolumn{10}{@{}l}{\textit{Vision--language / action models}} \\
Cosmos-Reason2 (video) & $-0.125$ & $0.027$ & $0.34$ & $-0.117$ & $0.012$ & $0.26$ & -- & -- & -- \\
GR00T N1.7 (2 frames) & $-0.087$ & $0.022$ & $0.37$ & $-0.129$ & $0.007$ & $0.25$ & $0.020$ & $0.040$ & $0.78$ \\
$\pi_0$ & $-0.213$ & $0.000$ & $0.02$ & $-0.098$ & $0.016$ & $0.29$ & $-0.014$ & $0.049$ & $1.13$ \\
OpenVLA-7B & $-0.080$ & $0.008$ & $0.24$ & $-0.061$ & $0.010$ & $0.31$ & -- & -- & -- \\
\midrule
V-JEPA 2.1 ViT-G + LoRA & -- & -- & -- & -- & -- & -- & $0.234$ & $0.253$ & $1.05$ \\
\bottomrule
\end{tabular}
\end{table*}

%% file: tables/strict_contact/table_loss_objective.tex
\begin{table}[tp]
\centering
\caption{Training objectives for the readout on final-layer features. Contact steps have a measured force of at least 2\,N, and $R^2_{\mathrm{strict}}$ uses only these steps. Our objective gives the highest contact $R^2$ for every model and view.}
\label{tab:loss_objective}
\footnotesize
\setlength{\tabcolsep}{3pt}
\begin{tabular}{@{}lcccc@{}}
\toprule
Objective & $r\uparrow$ & $R^2\uparrow$ & $A$ & $R^2_{\mathrm{strict}}\uparrow$ \\
\midrule
\multicolumn{5}{@{}l}{\textit{V-JEPA 2.1 ViT-G, gripper view}} \\
Smooth-$L_1$, all steps (ours) & $0.431$ & $0.085$ & $0.67$ & $-0.031$ \\
Squared error, all steps & $0.320$ & $-0.338$ & $0.90$ & $-0.308$ \\
Smooth-$L_1$, contact steps & $0.276$ & $-0.513$ & $0.69$ & $-0.312$ \\
Squared error, contact steps & $0.216$ & $-0.290$ & $1.19$ & $-0.284$ \\
\midrule
\multicolumn{5}{@{}l}{\textit{V-JEPA 2.1 ViT-G, human view}} \\
Smooth-$L_1$, all steps (ours) & $0.359$ & $-0.087$ & $1.00$ & $-0.230$ \\
Squared error, all steps & $0.304$ & $-0.318$ & $1.09$ & $-0.353$ \\
Smooth-$L_1$, contact steps & $0.153$ & $-0.205$ & $0.82$ & $-0.250$ \\
Squared error, contact steps & $0.127$ & $-1.025$ & $1.16$ & $-0.599$ \\
\midrule
\multicolumn{5}{@{}l}{\textit{VideoMAEv2 giant, human view}} \\
Smooth-$L_1$, all steps (ours) & $0.279$ & $-0.028$ & $0.36$ & $-0.169$ \\
Squared error, all steps & $0.108$ & $-0.380$ & $0.59$ & $-0.210$ \\
Smooth-$L_1$, contact steps & $0.120$ & $-0.500$ & $0.42$ & $-0.118$ \\
Squared error, contact steps & $0.102$ & $-2.073$ & $1.16$ & $-1.239$ \\
\midrule
\multicolumn{5}{@{}l}{\textit{DINOv3 ViT-H+, human view}} \\
Smooth-$L_1$, all steps (ours) & $0.044$ & $-0.036$ & $0.22$ & $-0.151$ \\
Squared error, all steps & $0.119$ & $-0.110$ & $0.25$ & $0.021$ \\
Smooth-$L_1$, contact steps & $0.096$ & $-0.421$ & $0.16$ & $-0.078$ \\
Squared error, contact steps & $0.105$ & $-1.017$ & $0.57$ & $-0.359$ \\
\midrule
\multicolumn{5}{@{}l}{\textit{VGGT-Omega 1B, human view}} \\
Smooth-$L_1$, all steps (ours) & $0.193$ & $-0.058$ & $0.49$ & $-0.208$ \\
Squared error, all steps & $0.084$ & $-0.270$ & $0.64$ & $-0.218$ \\
Smooth-$L_1$, contact steps & $0.052$ & $-0.387$ & $0.62$ & $-0.156$ \\
Squared error, contact steps & $0.046$ & $-1.294$ & $0.92$ & $-0.679$ \\
\bottomrule
\end{tabular}
\end{table}

%% file: tables/experiment_findings/figure_location_selection.tex
\begin{figure*}[p]
\centering
\includegraphics[width=\textwidth]{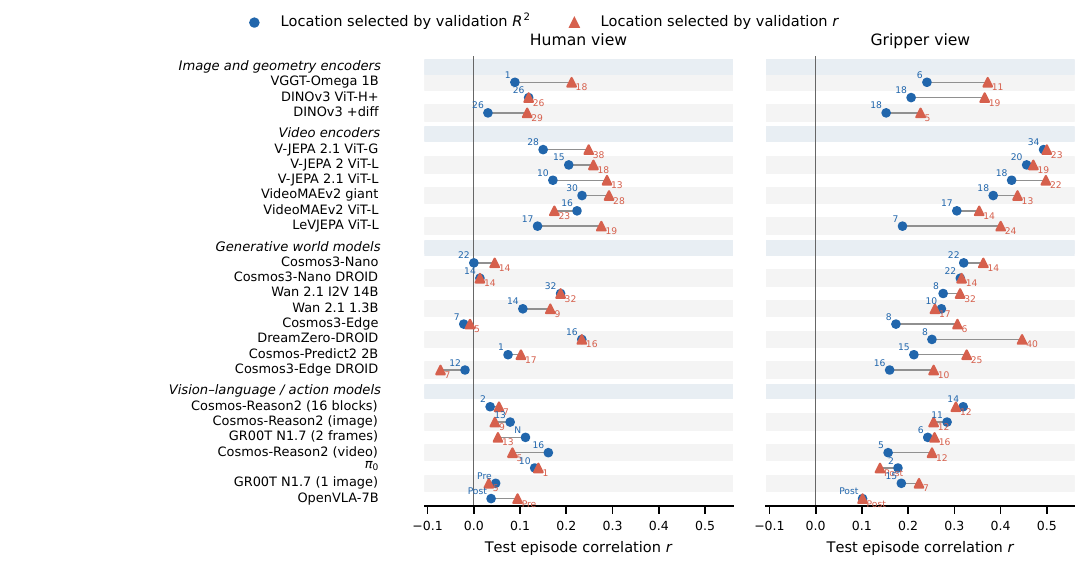}
\caption{Test correlation on \textbf{Hoi!} when the probe location is selected by validation $R^2$ (circles) or by validation $r$ (triangles). Labels give the selected block; Pre/Post mark the projector or merger and N the final normalization. Selecting by $r$ improves test correlation in 33 of 48 comparisons and lowers it in 10.}
\label{fig:location_selection}
\end{figure*}

%% file: tables/experiment_findings/table_selection_contact_r2.tex
\begin{table*}[tp]
\centering
\caption{Contact $R^2$ on \textbf{Hoi!} at the location selected by validation $R^2$ ($\ell^*$, original rule) and by validation $r$ ($\ell^*_r$). Contact $R^2$ is lower at $\ell^*_r$ for most models, which is why we report it at $\ell^*$.}
\label{tab:selection_contact_r2}
\small
\setlength{\tabcolsep}{9pt}
\begin{tabular}{@{}lrrrr@{}}
\toprule
& \multicolumn{2}{c}{Human contact $R^2$} & \multicolumn{2}{c}{Gripper contact $R^2$} \\
\cmidrule(lr){2-3}\cmidrule(lr){4-5}
Representation & Original rule & Validation-$r$ location & Original rule & Validation-$r$ location \\
\midrule
\multicolumn{5}{@{}l}{\textit{Image and geometry encoders}} \\
VGGT-Omega 1B & -0.047 & -0.129 & 0.018 & -0.057 \\
DINOv3 ViT-H+ & -0.033 & -0.033 & 0.008 & -0.077 \\
DINOv3 +diff & -0.105 & -0.102 & -0.022 & -0.104 \\
\midrule
\multicolumn{5}{@{}l}{\textit{Video encoders}} \\
V-JEPA 2.1 ViT-G & -0.055 & -0.131 & 0.166 & 0.092 \\
V-JEPA 2 ViT-L & -0.112 & -0.419 & 0.095 & 0.055 \\
V-JEPA 2.1 ViT-L & -0.019 & -0.152 & 0.110 & 0.157 \\
VideoMAEv2 giant & -0.037 & -0.047 & -0.012 & -0.065 \\
VideoMAEv2 ViT-L & -0.112 & -0.098 & -0.029 & -0.053 \\
LeVJEPA ViT-L & -0.079 & -0.017 & -0.069 & -0.052 \\
\midrule
\multicolumn{5}{@{}l}{\textit{Generative world models}} \\
Cosmos3-Nano & -0.297 & -0.251 & -0.037 & -0.074 \\
Cosmos3-Nano DROID & -0.263 & -0.263 & -0.063 & -0.083 \\
Wan 2.1 I2V 14B & -0.097 & -0.097 & -0.064 & -0.213 \\
Wan 2.1 1.3B & -0.177 & -0.254 & -0.047 & -0.162 \\
Cosmos3-Edge & -0.192 & -0.227 & -0.204 & -0.068 \\
DreamZero-DROID & -0.069 & -0.069 & -0.118 & -0.141 \\
Cosmos-Predict2 2B & -0.168 & -0.421 & -0.146 & -0.528 \\
Cosmos3-Edge DROID & -0.215 & -0.128 & -0.086 & -0.045 \\
\midrule
\multicolumn{5}{@{}l}{\textit{Vision--language / action models}} \\
Cosmos-Reason2 (16 blocks) & -0.120 & -0.064 & -0.028 & -0.083 \\
Cosmos-Reason2 (image) & -0.111 & -0.120 & -0.031 & -0.074 \\
GR00T N1.7 (2 frames) & -0.087 & -0.113 & -0.129 & -0.103 \\
Cosmos-Reason2 (video) & -0.125 & -0.113 & -0.117 & -0.223 \\
$\pi_0$ & -0.213 & -0.036 & -0.098 & -0.166 \\
GR00T N1.7 (1 image) & -0.124 & -0.039 & -0.139 & -0.079 \\
OpenVLA-7B & -0.080 & -0.394 & -0.061 & -0.061 \\
\bottomrule
\end{tabular}
\end{table*}

%% file: tables/attentive_robustness/table_attentive_robustness.tex
\begin{table*}[tp]
\centering
\caption{Attentive readout on the final-layer human-view features of \textbf{Hoi!}. Affine/max: our probe. Attentive: mean $\pm$ SD over three seeds. The attentive probe helps V-JEPA but does not change the ranking.}
\label{tab:attentive_robustness}
\small
\setlength{\tabcolsep}{6pt}
\begin{tabular}{@{}lrrrr@{}}
\toprule
& \multicolumn{2}{c}{Episode correlation $r\uparrow$} & \multicolumn{2}{c}{Contact $R^2\uparrow$} \\
\cmidrule(lr){2-3}\cmidrule(lr){4-5}
Representation & Affine/max & Attentive & Affine/max & Attentive \\
\midrule
\multicolumn{5}{@{}l}{\textit{Image and geometry encoders}} \\
DINOv3 ViT-H+ & 0.108 & $0.170\,\pm\,0.029$ & -0.060 & $-0.632\,\pm\,0.068$ \\
VGGT-Omega 1B & 0.215 & $0.149\,\pm\,0.020$ & -0.087 & $-0.086\,\pm\,0.005$ \\
DINOv3 +diff & 0.072 & $0.143\,\pm\,0.030$ & -0.067 & $-0.814\,\pm\,0.252$ \\
\midrule
\multicolumn{5}{@{}l}{\textit{Video encoders}} \\
V-JEPA 2.1 ViT-G & 0.368 & $0.443\,\pm\,0.024$ & -0.144 & $0.070\,\pm\,0.035$ \\
VideoMAEv2 giant & 0.313 & $0.259\,\pm\,0.018$ & -0.071 & $-0.189\,\pm\,0.069$ \\
\bottomrule
\end{tabular}
\end{table*}

%% file: figures/qualitative/supp/figures_supp_overlays.tex
\begin{figure*}[p]
\centering
\includegraphics[width=\textwidth]{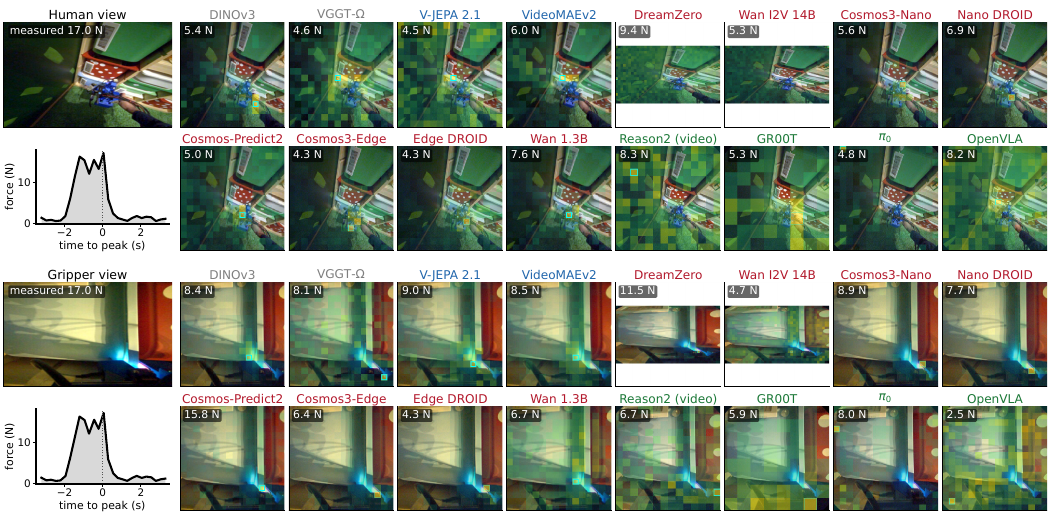}\\[6pt]
\includegraphics[width=\textwidth]{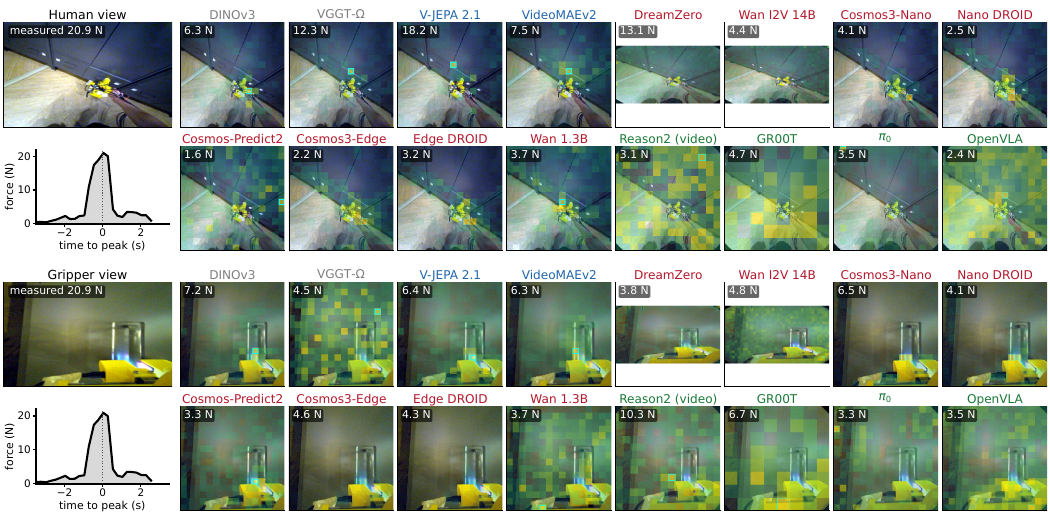}
\caption{Readout maps of all models on \textbf{Hoi!} at the force peak of the contacts in Fig.~3a (top) and Fig.~3b (bottom) of the main paper. For each contact, we show the human view (upper rows) and the gripper view (lower rows) at the same instant. Each tile shows the affine map before max pooling on the model's input frame, with brighter tokens scoring higher, the token selected by max pooling, and the predicted force. Title colors mark the model family.}
\label{fig:supp_overlay_hoi}
\end{figure*}

\begin{figure*}[t]
\centering
\includegraphics[width=\textwidth]{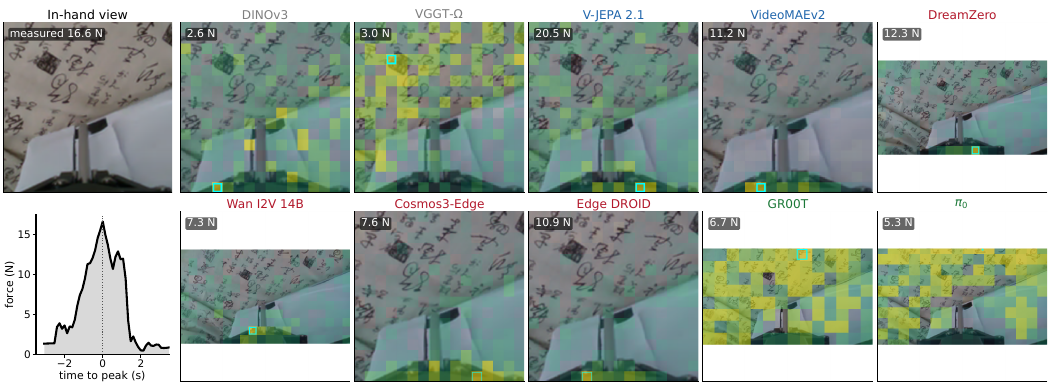}\\[6pt]
\includegraphics[width=\textwidth]{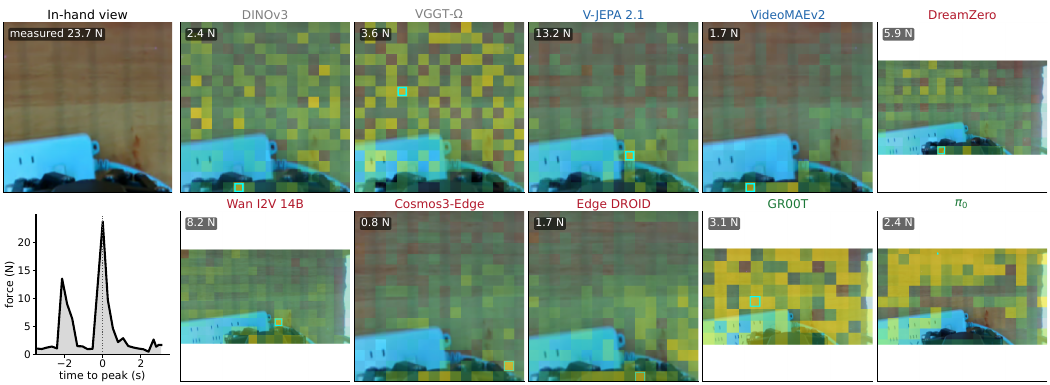}
\caption{Readout maps of all models on the two \textbf{RH20T} recordings of Fig.~4 of the main paper at the force peak: (top) opening a microwave door and (bottom) pressing the button of a power strip. Layout as in Fig.~\ref{fig:supp_overlay_hoi}.}
\label{fig:supp_overlay_rh20t}
\end{figure*}

%% file: figures/probe_locations/supplementary_figures.tex
\clearpage
\begin{figure*}[p]
\centering
\includegraphics[width=\linewidth,height=.78\textheight,keepaspectratio]{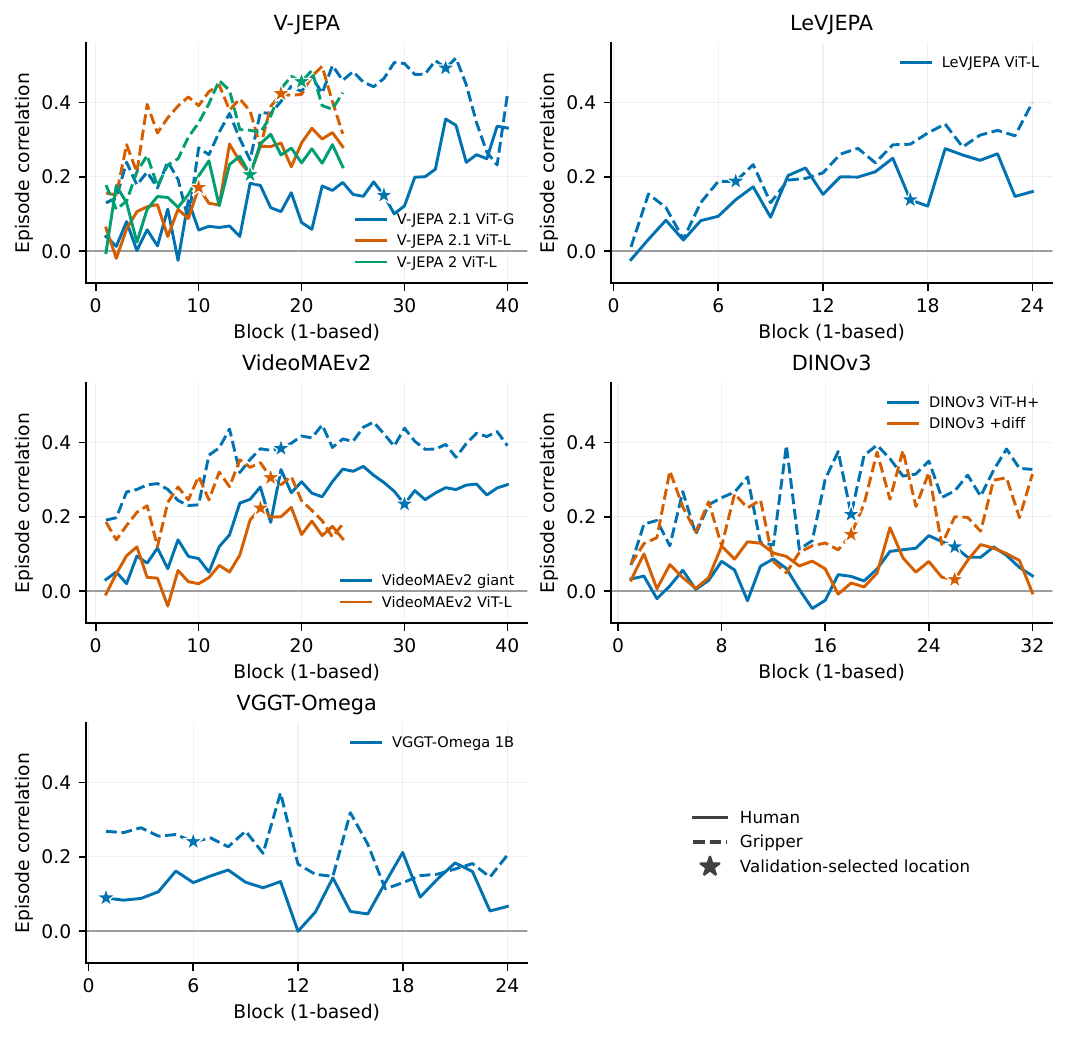}
\caption{Force tracking ($r$) across depth for the image and video encoders on the \textbf{Hoi!} test episodes. Stars mark the locations selected on validation.}
\label{fig:encoders_test_r}
\end{figure*}
\clearpage
\begin{figure*}[p]
\centering
\includegraphics[width=\linewidth,height=.78\textheight,keepaspectratio]{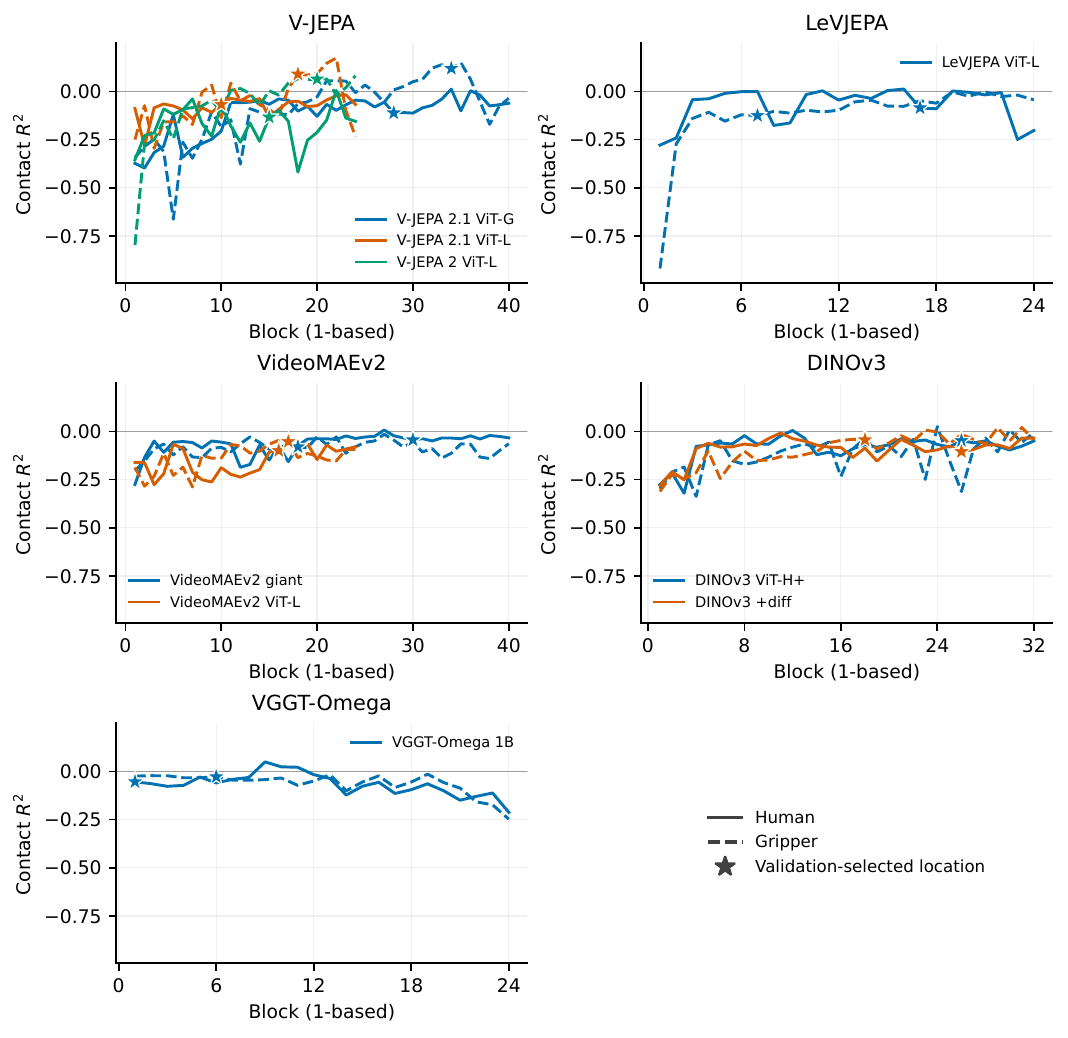}
\caption{Contact $R^2$ across depth for the image and video encoders on the \textbf{Hoi!} test episodes. Layout as in Fig.~\ref{fig:encoders_test_r}.}
\label{fig:encoders_test_r2}
\end{figure*}
\clearpage
\begin{figure*}[p]
\centering
\includegraphics[width=\linewidth,height=.78\textheight,keepaspectratio]{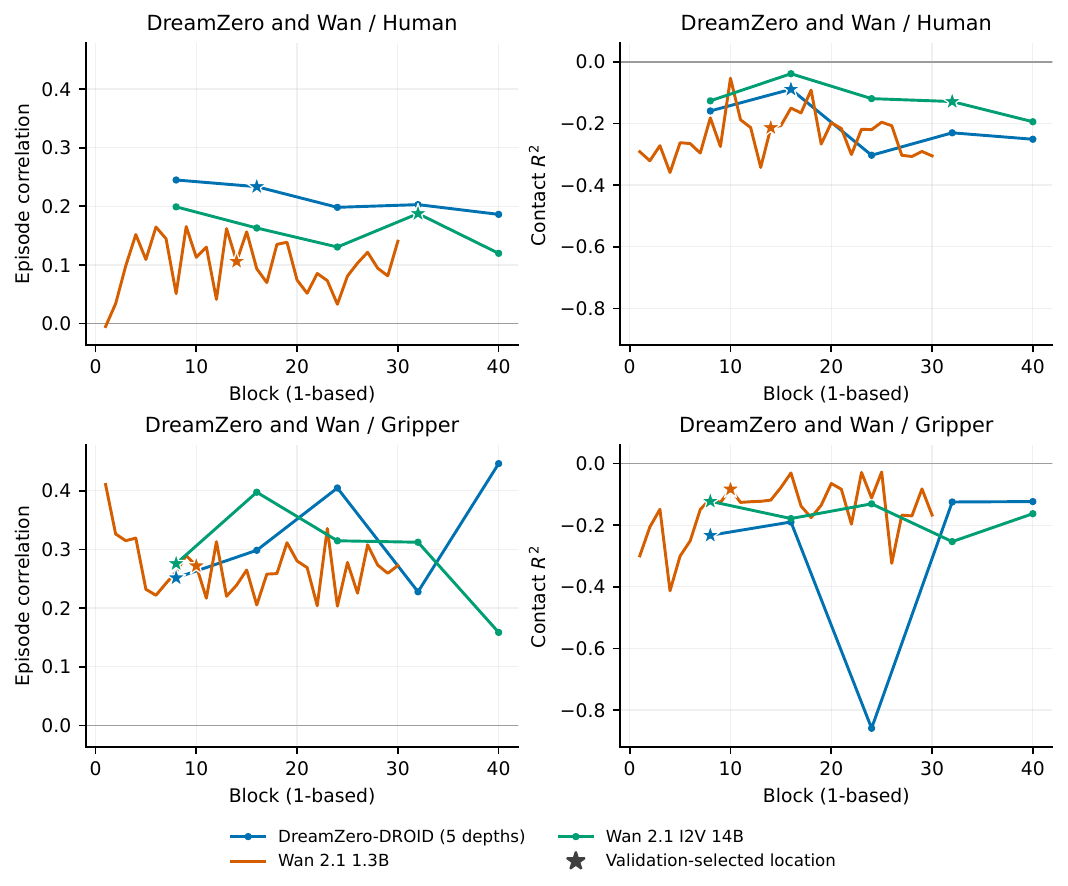}
\caption{Force tracking and contact $R^2$ across depth for Wan~2.1~1.3B, Wan I2V~14B, and DreamZero on the \textbf{Hoi!} test episodes. Wan I2V~14B and DreamZero are probed at five depths.}
\label{fig:world_wan_dreamzero}
\end{figure*}
\clearpage
\begin{figure*}[p]
\centering
\includegraphics[width=\linewidth,height=.78\textheight,keepaspectratio]{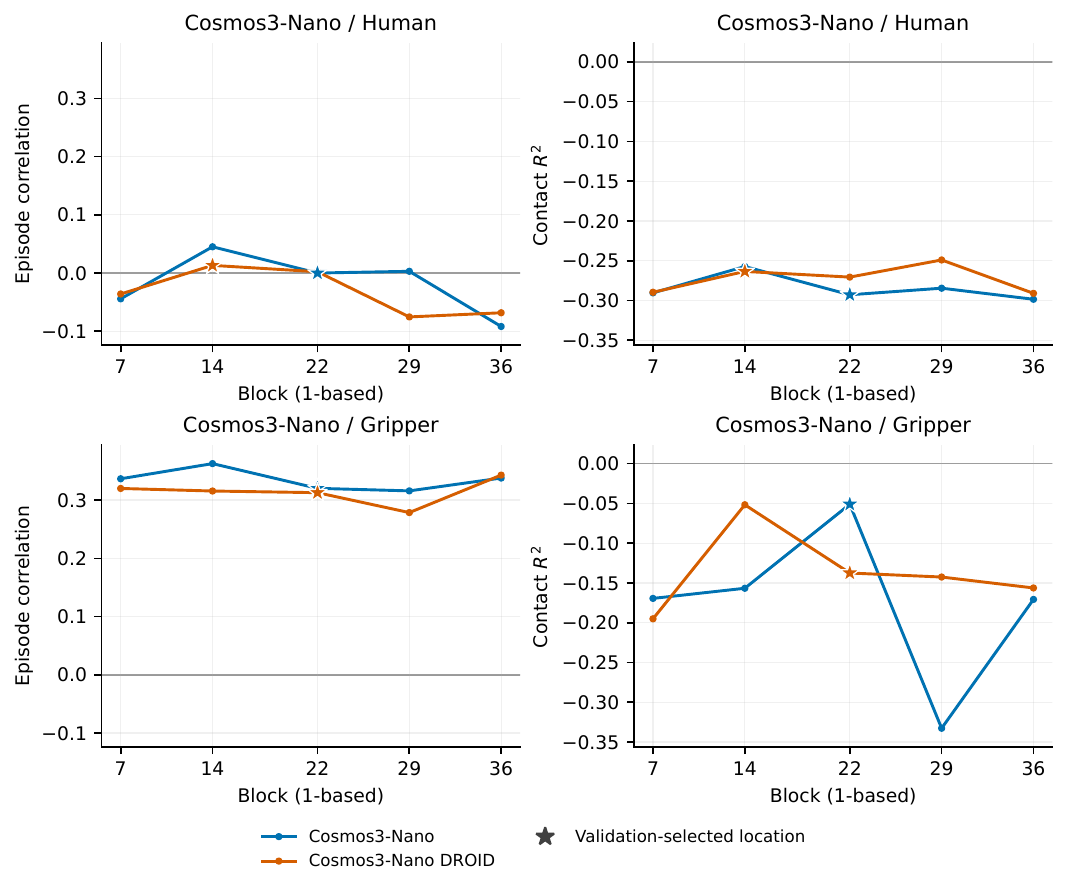}
\caption{Force tracking and contact $R^2$ across depth for Cosmos3-Nano and Nano DROID on the \textbf{Hoi!} test episodes, probed at five depths.}
\label{fig:world_cosmos3_nano}
\end{figure*}
\clearpage
\begin{figure*}[p]
\centering
\includegraphics[width=\linewidth,height=.78\textheight,keepaspectratio]{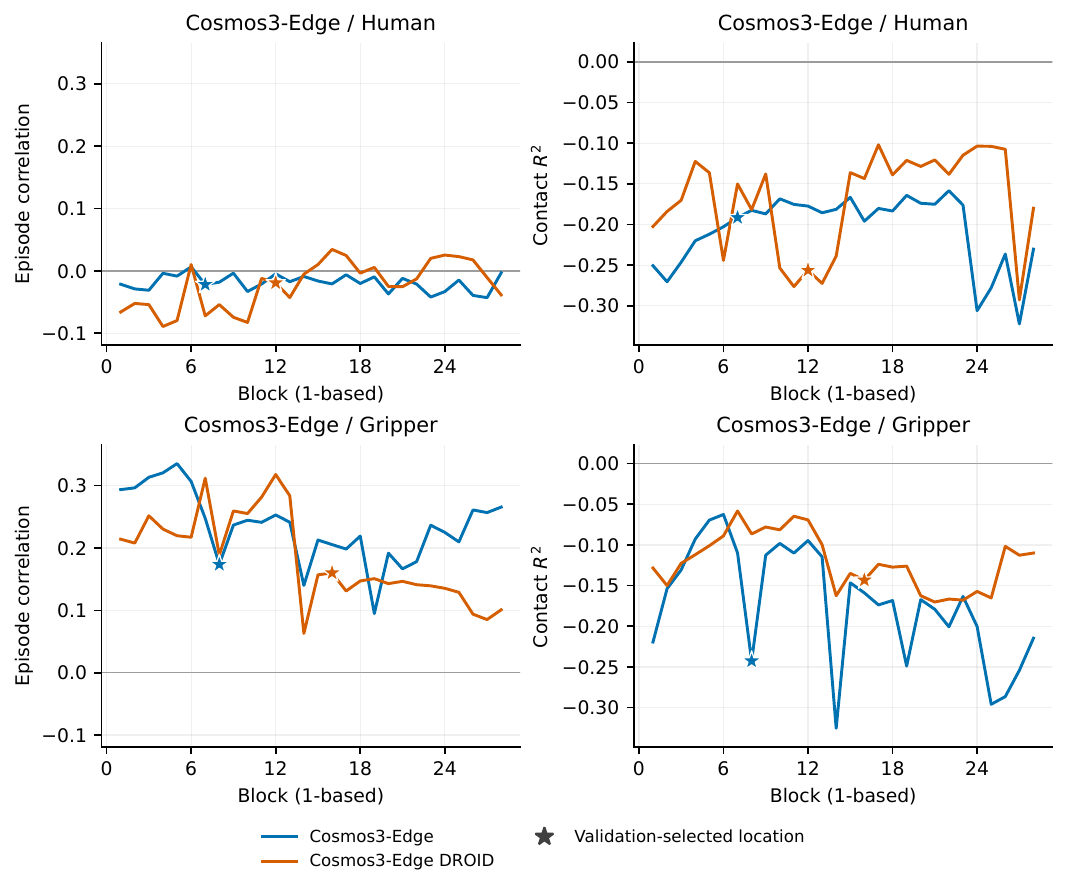}
\caption{Force tracking and contact $R^2$ across depth for Cosmos3-Edge and Edge DROID on the \textbf{Hoi!} test episodes.}
\label{fig:world_cosmos3_edge}
\end{figure*}
\clearpage
\begin{figure*}[p]
\centering
\includegraphics[width=\linewidth,height=.78\textheight,keepaspectratio]{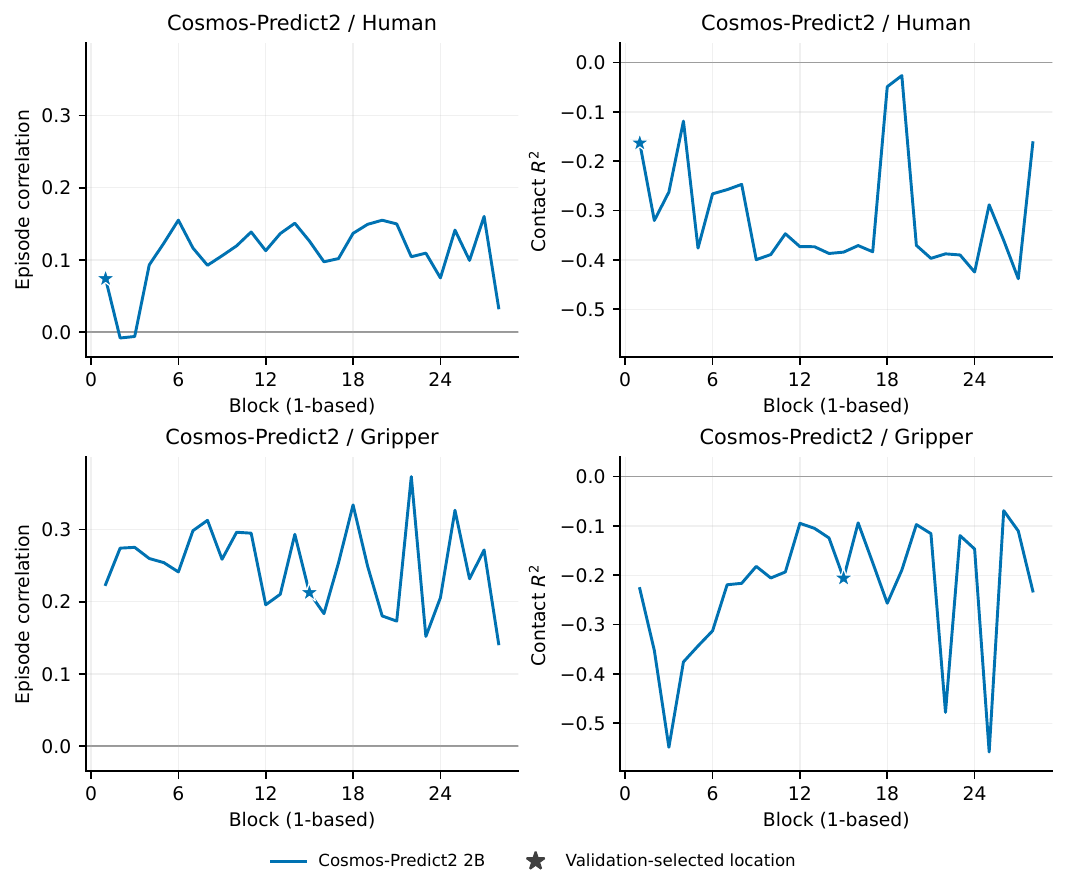}
\caption{Force tracking and contact $R^2$ across depth for Cosmos-Predict2 on the \textbf{Hoi!} test episodes.}
\label{fig:world_cosmos_predict2}
\end{figure*}
\clearpage
\begin{figure*}[p]
\centering
\includegraphics[width=\linewidth,height=.78\textheight,keepaspectratio]{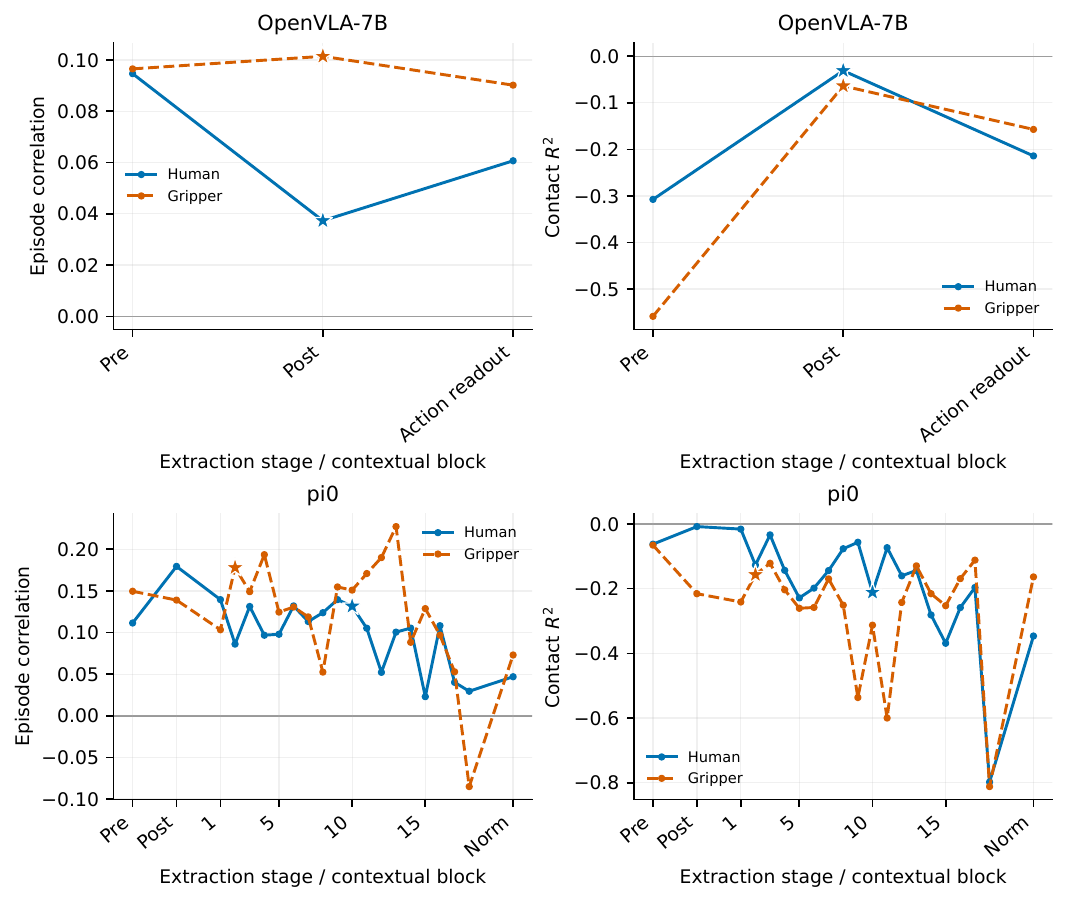}
\caption{Force tracking and contact $R^2$ across the extraction stages of OpenVLA and $\pi_0$ on the \textbf{Hoi!} test episodes. Pre/Post: before and after the projection into the language model. Numbers: language-model blocks. Norm: final normalization.}
\label{fig:policy_other_stages}
\end{figure*}
\clearpage
\begin{figure*}[p]
\centering
\includegraphics[width=\linewidth,height=.78\textheight,keepaspectratio]{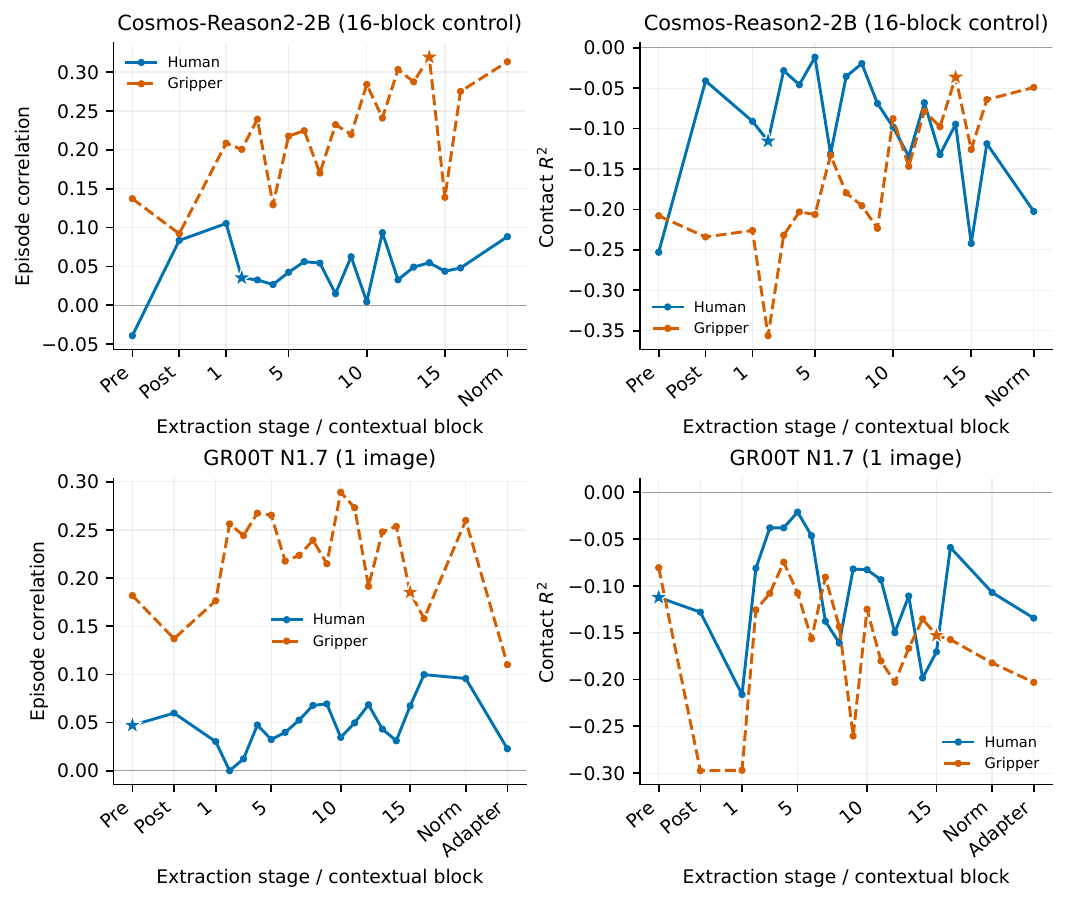}
\caption{Force tracking and contact $R^2$ across the extraction stages of GR00T~N1.7 and the matched 16-block Cosmos-Reason2 with a single image per time step, on the \textbf{Hoi!} test episodes. Pre/Post: before and after the vision merger. Layout as in Fig.~\ref{fig:policy_other_stages}.}
\label{fig:policy_groot_reason2_image}
\end{figure*}
\clearpage
\begin{figure*}[p]
\centering
\includegraphics[width=\linewidth,height=.78\textheight,keepaspectratio]{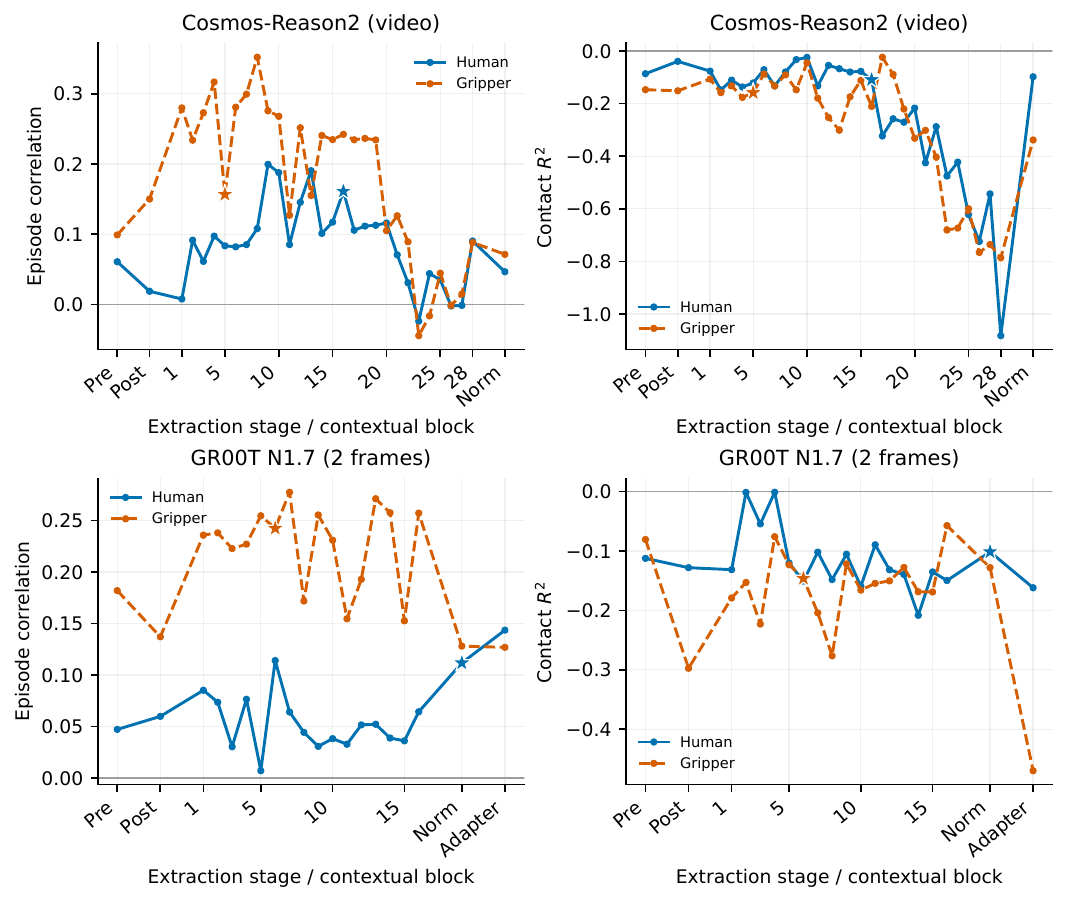}
\caption{Force tracking and contact $R^2$ across the extraction stages of Cosmos-Reason2 with video input and GR00T~N1.7 with its native two-image input, on the \textbf{Hoi!} test episodes. Layout as in Fig.~\ref{fig:policy_groot_reason2_image}.}
\label{fig:policy_groot_reason2_temporal}
\end{figure*}
\clearpage
\begin{figure*}[p]
\centering
\includegraphics[width=\linewidth,height=.78\textheight,keepaspectratio]{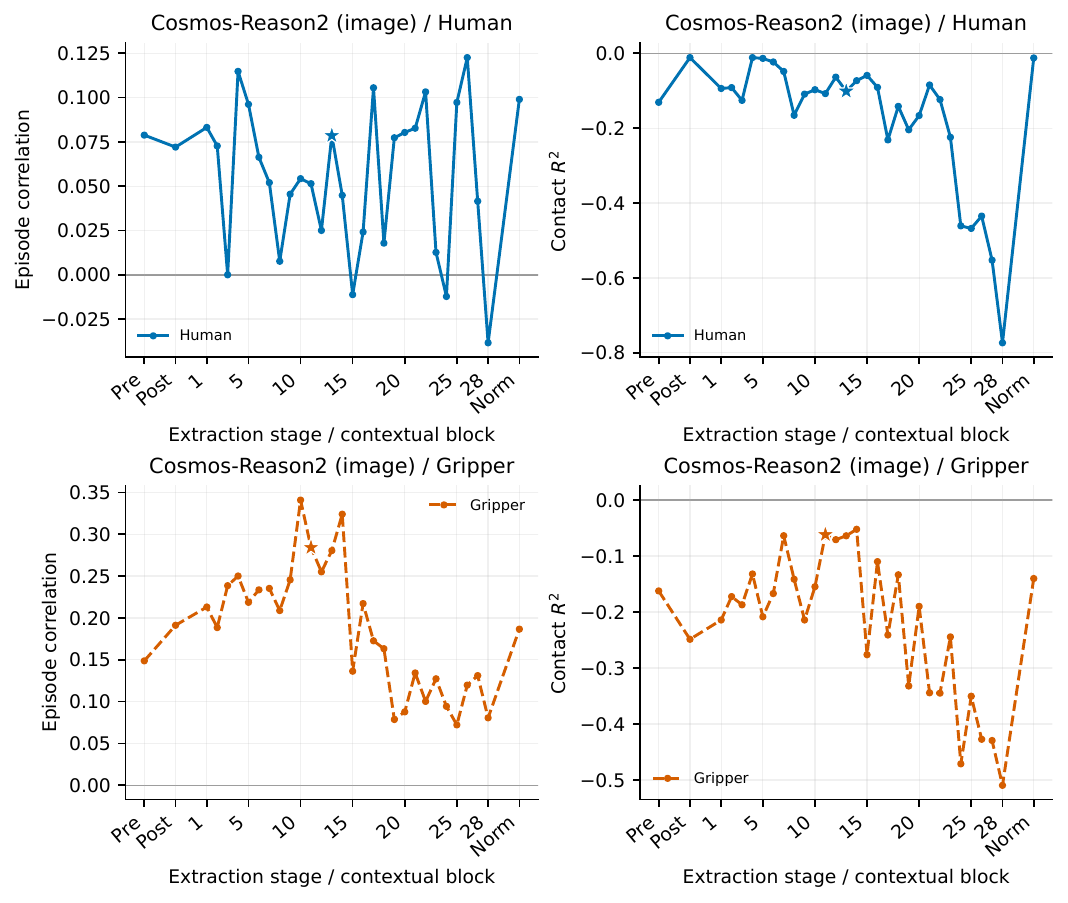}
\caption{Force tracking and contact $R^2$ across the extraction stages of the full 28-block Cosmos-Reason2 with a single image per time step, on the \textbf{Hoi!} test episodes. Layout as in Fig.~\ref{fig:policy_groot_reason2_image}.}
\label{fig:policy_reason2_full_image}
\end{figure*}
\clearpage
\begin{figure*}[p]
\centering
\includegraphics[width=\linewidth,height=.78\textheight,keepaspectratio]{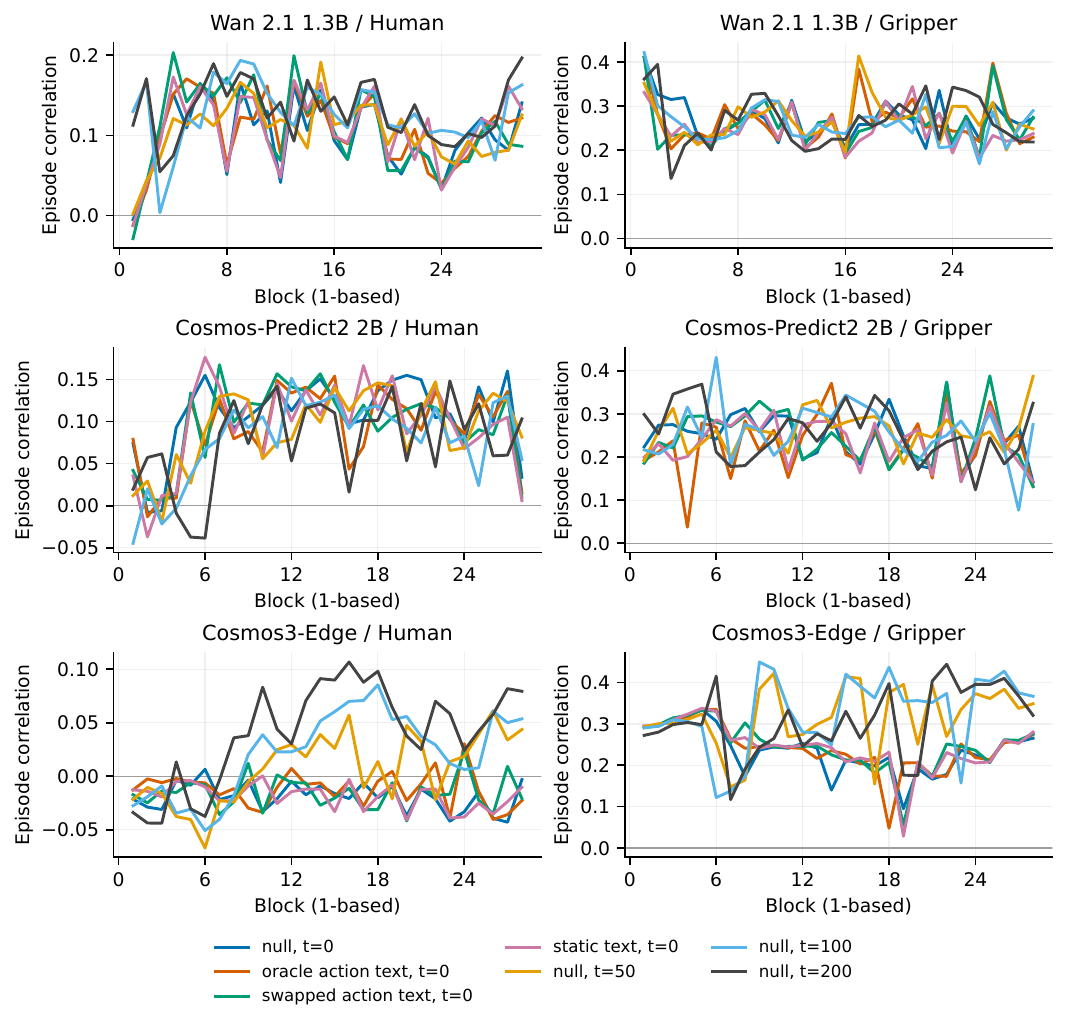}
\caption{Force tracking across world-model depth on the \textbf{Hoi!} test episodes for every text prompt and noise level (Sec.~\ref{sec:supp_depth}). Only the empty-prompt, noise-free condition enters the main tables.}
\label{fig:world_conditions}
\end{figure*}